\documentclass{article} 
\usepackage{iclr2027_conference,times}

\usepackage{amsmath,amsfonts,bm}

\def\eqref#1{equation~\ref{#1}}

\def\1{\bm{1}}

\DeclareMathAlphabet{\mathsfit}{\encodingdefault}{\sfdefault}{m}{sl}
\SetMathAlphabet{\mathsfit}{bold}{\encodingdefault}{\sfdefault}{bx}{n}

\usepackage{hyperref}
\usepackage{url}

\title{Does On-Policy Distillation Really Distill? From Noisy Teacher to Self-Improvement}

\author{Yi Ding, Ruqi Zhang \\
Department of Computer Science,
Purdue University,
USA \\
\texttt{\{ding432,ruqiz\}@purdue.edu}
}

\usepackage{xcolor}
\usepackage[table]{xcolor}

\definecolor{mygreen}{HTML}{298D66}
\definecolor{myred}{HTML}{C62828}

\definecolor{cite}{HTML}{8491B4}
\definecolor{link}{HTML}{669999}

\hypersetup{
    colorlinks=true,
    citecolor=cite,  
    linkcolor=myred,  
    urlcolor=link,
}

\usepackage{amsthm}

\theoremstyle{definition}

\usepackage{booktabs}
\usepackage{multirow}
\usepackage{graphicx}
\usepackage{pifont}
\usepackage{wrapfig}
\usepackage{enumitem}
\usepackage{fontawesome}
\usepackage{twemojis}
\usepackage[most,skins,theorems]{tcolorbox}
\tcbset{
  takeaway/.style={
    width=\linewidth,
    top=12pt,
    bottom=8pt,
    left=-12pt,
    right=12pt,
    colback=cite!20,
    colframe=black,
    colbacktitle=cite!50!black,
    enhanced,
    breakable,
    center,
    fontupper=\normalsize,
    attach boxed title to top left={yshift=-0.1in,xshift=0.15in},
    boxed title style={boxrule=0pt,colframe=white,},
  }
}

\newtcolorbox{takeaway}[2][]{takeaway,title=#2,#1}

\tcbset{
  qs/.style={
    width=\linewidth,
    top=10pt,
    bottom=6pt,
    left=12pt,
    right=12pt,
    colback=cite!20,
    colframe=black,
    colbacktitle=cite!50!black,
    enhanced,
    breakable,
    center,
    fontupper=\normalsize,
    attach boxed title to top left={yshift=-0.1in,xshift=0.15in},
    boxed title style={boxrule=0pt,colframe=white,},
  }
}

\newtcolorbox{question}[2][]{qs,title=#2,#1}

\iclrfinalcopy 
\begin{document}

\maketitle

\begin{center}
  \vspace{-30pt}
  \href{https://huggingface.co/collections/Tuwhy/on-policy-self-adaptation}{\texttwemoji{1f917}\;\textcolor{cite}{Hugging Face}}
  \quad\quad
  \href{https://github.com/DripNowhy/On-Policy-Self-Adaptation}{\textcolor{black}{\faGithub\;}\textcolor{cite}{Github}}
\end{center}

\begin{abstract}

On-policy distillation (OPD) offers dense token-level supervision as an alternative to the sparse outcome-level advantages of reinforcement learning with verifiable rewards (RLVR). 
However, the teacher scores student-generated trajectories that are inherently off-policy for it, so the reliability of its supervision, and hence the source of the student's improvement, remains unclear.
We quantitatively analyze teacher supervision during OPD training and find substantial noise whose prevalence increases with teacher scale. Surprisingly, the student policy is insensitive to such noise, converging to comparable performance regardless of whether noisy supervision is retained or removed.
Does OPD distill at all? By analyzing what drives its gains, we find that learning concentrates on low log-probability tokens, and using a single fixed negative advantage matches the performance of teacher-provided ones. This suggests that OPD works largely by suppressing low log-probability tokens, which requires no teacher. These findings motivate On-Policy Self-Adaptation (OPSA), a supervision-free method using entropy-adaptive negative advantages.
It assigns stronger learning signals to high-entropy positions, suppressing tail tokens, and evenly redistributing probability mass among head tokens.
Compared with the base \texttt{Qwen3-1.7B}, OPSA improves Avg@32 by 35.41 points on AIME24, corresponding to a 263\% relative gain, and more than doubles Pass@32 across all three benchmarks. It also outperforms OPD by 16.77 points in Avg@32 on AIME24.
Extensive experiments and analyses across model families and tasks further demonstrate its effectiveness and generalizability.

\end{abstract}

\begin{figure}[h]
    \centering
    \includegraphics[width=1\linewidth]{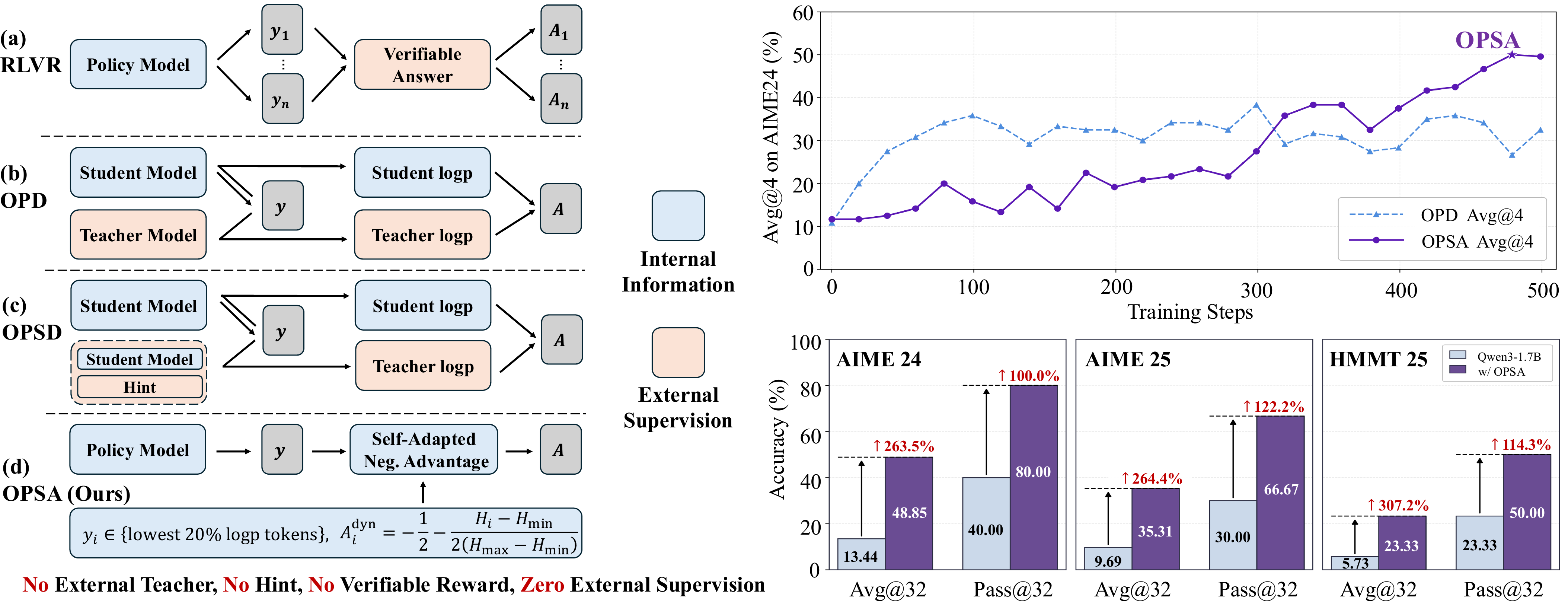}
    \caption{\textbf{Left:} Overview of different on-policy reinforcement learning algorithms. Unlike existing methods that derive advantages from external supervision, OPSA enables self-improvement with no external supervision by assigning entropy-adaptive negative advantages to low-probability tokens.  This suppresses tail tokens and redistributes their mass among head tokens, sharpening low-entropy positions while preserving diversity at high-entropy forks. \textbf{Right:} Training dynamics and performance of OPSA. OPSA eliminates the teacher supervision used in OPD, and further outperforms OPD by 12\% in Avg@4. On \texttt{Qwen3-1.7B}, OPSA improves Avg@32 by 263-307\% across mathematical reasoning benchmarks and more than doubles Pass@32.}
    \label{fig:overview}
\end{figure}

\section{Introduction}

Reinforcement learning (RL) has become a dominant post-training paradigm to improve reasoning capabilities of large language models (LLMs) ~\citep{guo2025deepseek,yang2025qwen3,glm5team2026glm5vibecodingagentic}, offering stronger performance and generalization than offline methods such as supervised fine-tuning (SFT)~\citep{chu2025sft,chen2025sft}. Reinforcement learning with verifiable rewards (RLVR), including GRPO~\citep{shao2024deepseekmath} and DAPO~\citep{yu2026dapo}, samples multiple responses per question and assigns advantages by normalizing verifiable rewards within each group (Fig.~\ref{fig:overview}(a)). However, response-level rewards provide coarse and sparse supervision for long-horizon reasoning~\citep{yue2025vapo}. Moreover, when responses within a group share the same correctness outcome, their normalized advantages vanish~\citep{wang2026vl,dinglearning}, weakening the learning signal and making training unstable or prone to collapse.

On-Policy Distillation (OPD)~\citep{agarwal2024policy,gu2024minillm} addresses this limitation by using a strong teacher to provide token-level advantages through reverse Kullback-Leibler (KL) optimization on student-sampled trajectories (Fig.~\ref{fig:overview}(b))~\citep{burnham2001kullback,lu2025onpolicydistillation}. However, it requires shared vocabularies and white-box access to teacher logits, restricting teacher selection. On-Policy Self-Distillation (OPSD)~\citep{hubotter2026reinforcement,zhao2026self,shenfeld2026self} retains the same OPD paradigm but replaces the external teacher with the policy itself conditioned on hints, such as reference answers (Fig.~\ref{fig:overview}(c)). Constructing such hints still requires additional sampling or annotation and may risk information leakage~\citep{yang2026self}.

More fundamentally, OPD and OPSD share the same advantage-assignment paradigm, which asks the teacher to score student-sampled prefixes that are inherently off-policy for it. This raises the question of whether the teacher can still provide reliable supervision in such off-policy contexts~\citep{xie2026position,fu2026revisiting,hou2026uni}.
Our analysis reveals a substantial fraction of noisy teacher supervision, where \emph{noisy} means a negative advantage on a correct answer or a positive advantage on an incorrect one, and this fraction grows with teacher scale. Surprisingly, training exclusively on these noisy trajectories achieves performance comparable to both standard OPD and OPD trained with the noisy trajectories removed, suggesting that OPD's gains may not arise from the behavior matching it is designed to achieve. This raises a fundamental question: \emph{what actually drives student improvement in OPD?}

To answer this question, we analyze OPD’s gains from the perspectives of token selection and learning signals. We find that (i) improvement is driven primarily by low log-probability tokens sampled by the student, and (ii) negative advantages are critical, as replacing all OPD advantages with a single fixed negative value yields comparable performance. These observations suggest that OPD’s gains come from suppressing low-probability tokens sampled by the student, requiring no teacher. 
Fixed negative advantages reproduce much of OPD’s gains, but they discard its fine-grained token-level signals by treating all tokens equally. We therefore study how much negative signal each token should receive, and find that token entropy determines it, with stronger signal at high-entropy positions giving better performance.

Altogether, we propose On-Policy Self-Adaptation (OPSA), a supervision-free token-level RL method that assigns entropy-adaptive negative advantages to low-probability tokens (Fig.~\ref{fig:overview}(d)). 
We summarize our contributions as follows:

\begin{itemize}
\item We find that teacher supervision in OPD is highly noisy and uncover a surprising insensitivity of the student policy to such noise. Training exclusively with noisy supervision, excluding noisy supervision, and standard OPD all converge to comparable accuracy after similar numbers of training steps.
\item We identify the key drivers of OPD improvement: effective learning is concentrated on low-logp student-sampled tokens, and fixed negative advantages on these tokens can match standard OPD. This suggests that OPD’s gains stem less from teacher distillation than from suppressing low-probability tokens, questioning the necessity of teacher supervision.
\item We propose On-Policy Self-Adaptation, a supervision-free token-level RL method. It reshapes the policy distribution by suppressing tail-token probabilities and redistributing the probability mass among head tokens. This sharpens the distribution at low-entropy positions while preserving diversity at high-entropy fork tokens to support effective exploration. 
\item We show through extensive experiments that OPSA generalizes across model families and tasks. On \texttt{Qwen3-1.7B}, OPSA improves Avg@32 by 263\%--307\% relative to the base model across the three benchmarks and more than doubles Pass@32 on each benchmark.

\end{itemize}

\section{Teacher Supervision in OPD Is Highly Noisy, but Students Improve Regardless} 
\subsection{Preliminary}
On-Policy Distillation (OPD)~\citep{agarwal2024policy,gu2024minillm} optimizes the reverse Kullback–Leibler (KL) divergence between the student and teacher distributions over the same prefixes, which is computed using the K1 estimator~\citep{lu2025onpolicydistillation}:
\begin{equation}
    \mathrm{KL}(\pi_s\mid\mid\pi_t)=\mathbb{E}_{y\sim\pi_s(\cdot\mid x)}\sum_{i=1}^{\vert y\vert}\left[\log\pi_s(y_i\mid x;y_{<i})-\log\pi_t(y_i\mid x;y_{<i})\right],
\end{equation}
where $y_{<i}=[y_1,\cdots,y_{i-1}]$ denotes prefixes of the response $y$ sampled from student policy $\pi_s$. In practice, OPD provides token-level advantage signals $A_i$ for the student policy during the Reinforcement Learning (RL) training, and the loss function can be written as:
\begin{equation}\label{eq:opd_loss}
    \mathcal{L}_{\text{OPD}}=-\mathbb{E}\left[\frac{1}{\vert y\vert}\sum_{i=1}^{\vert y\vert}A_i\log\pi_s(y_i\vert x;y_{<i})\right], \hspace{10pt}A_i=\log\frac{\pi_t(y_i\mid x;y_{<i})}{\pi_s(y_i\mid x;y_{<i})}.
\end{equation}
The loss function optimizes the student policy by assigning positive or negative credits to the student-sampled tokens based on teachers' preference, pushing it closer to teachers' behavior in each state.

\subsection{Noisy Teacher Signals in OPD}\label{sec:2.2}
Since the KL divergence is computed conditioned on prefixes sampled from the student distribution, the teacher must score trajectories it would not generate. This raises a key concern: can the teacher provide reliable supervision in such off-policy settings~\citep{fu2026revisiting,xie2026position,wang2026backtracking}?

\paragraph{Definition of Noisy Signals.}
Previous work~\citep{hou2026uni} often measures supervision noise at the trajectory level. However, the quality of intermediate reasoning steps is difficult to assess quantitatively, as a per-token reference for what the supervision should be is generally unavailable. Therefore,
we focus on tokens corresponding to verifiable final answers enclosed in \verb|\boxed{}|, whose correctness is determined by the verifier. We call the supervision \emph{noisy} when the sign of the teacher-provided advantage on these tokens disagrees with the verifiable reward, i.e., when incorrect answers receive positive advantages or correct answers receive negative ones. Such supervision provides the student with misleading learning signals during RL training.

\paragraph{Setup.}
We use \texttt{Qwen3-1.7B} as the student policy $\pi_s$ with thinking mode disabled, and use \texttt{Qwen3-4B/30B-A3B/235B-A22B-Instruct} as teacher $\pi_t$, respectively.
For the following analysis, we randomly sample one correct and one incorrect response from $\pi_s$ for each of 500 questions in DAPO-17k~\citep{yu2026dapo}, with correctness verified based on ground-truth answers.

\begin{figure}[t]
    \centering
    \includegraphics[width=1.0\linewidth]{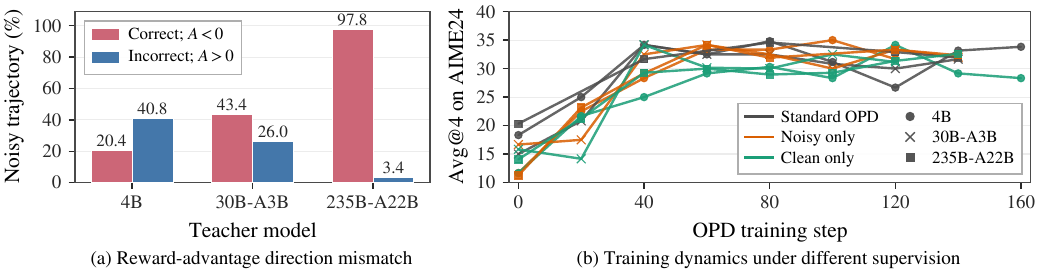}
    \caption{Performance of Qwen3-1.7B as the student model under supervision from teachers of different scales. (a): The proportion of teacher-provided advantage signals whose directions disagree with verifiable rewards on student-sampled trajectories. Red and blue bars denote noise rates in correct and incorrect trajectories, respectively. (b): Avg@4 performance on AIME24 during OPD training with different variants of teacher-provided advantage signals.}
    \label{fig:noisy_resilience}
\end{figure}

\paragraph{Teacher Supervision is Highly Noisy, and Noisier at Scale.}
As shown in Fig.~\ref{fig:noisy_resilience}(a), supervision from teacher models of all sizes contains substantial noise. For the 4B teacher, 20.4\% of correct trajectories receive negative advantages on their answer tokens, while 40.8\% of incorrect trajectories receive positive advantages, yielding an overall noise rate of 30.6\%. The noise rate increases with teacher size, from 30.6\% for the 4B teacher to 34.7\% for the 30B-A3B teacher and 50.6\% for the 235B-A22B teacher.
In addition, the largest teacher consistently tends to assign negative advantages: 97.8\% of answer tokens enclosed in \verb|\boxed{}| receive negative advantages even when the answer is correct, while 96.6\% also receive negative advantages when the answer is incorrect. This leads to an overall noise rate of approximately 50\%. These results suggest that, as the teacher becomes more capable, its supervision on student-generated trajectories becomes overwhelmingly negative and increasingly insensitive to answer correctness.
We attribute this to the growing distributional mismatch between the student and teacher policies, which makes student-generated trajectories increasingly off-policy from the teacher's perspective.

\subsection{Students Are Insensitive to Noise in Teacher Supervision}\label{sec:2.3}
Analysis in Section~\ref{sec:2.2} reveals that OPD training involves a substantial amount of noisy token-level supervision. To isolate its effect on learning, we conduct a controlled filtering experiment by partitioning trajectories according to whether they contain noisy signals defined in Section~\ref{sec:2.2}. We compare standard OPD trained on all trajectories against two variants trained exclusively on trajectories with or without noisy signals.
As shown in Fig.~\ref{fig:noisy_resilience}(b), all three settings converge to comparable performance after a similar number of gradient steps. Remarkably, even when training is restricted to trajectories containing noisy advantages, the student improves at a rate comparable to standard OPD. 
This finding reveals a surprising insensitivity of OPD to supervision noise in the teacher-provided advantages. 
In particular, when the teacher supervision is substantially noisy, the student’s improvement may not come from matching the teacher’s behavior. This raises a more fundamental question:

\begin{question}{Research Question}

\emph{OPD improves student performance even when teacher supervision is highly noisy. 
This suggests that the gains may not come from knowledge transfer, the mechanism OPD is built on. What, then, drives student improvement in OPD?}

\end{question}

\section{Where Does Student Improvement Come From?}\label{sec:3}
In this section, we progressively disentangle where the performance gains of OPD come from. Specifically, Sections~\ref{sec:3.1} and~\ref{sec:3.2} investigate which tokens affect improvement and what kinds of learning signals are effective, respectively. We summarize our findings as follows:

\begin{takeaway}{Key Findings}
    \begin{itemize}
    \item \textbf{Which Tokens?} Most high-log-probability tokens sampled by the student provide negligible gradients and little effective learning signal. OPD’s performance gains instead arise primarily from the small fraction of low-log-probability tokens.~(\S\ref{sec:3.1})
    \item \textbf{Which Signals?} Negative advantages, which constitute the majority of OPD signals, are the ones that matter. Notably, replacing all OPD advantages, both positive and negative, with a fixed negative value yields performance comparable to standard OPD.~(\S\ref{sec:3.2})
    \end{itemize}
\end{takeaway}

\subsection{Which Tokens Contribute to OPD Training?}\label{sec:3.1}

Findings in Section~\ref{sec:2.3} suggest that not all tokens generated by the student policy contribute effective learning signals for OPD. Revisiting the loss in Eq.~\ref{eq:opd_loss}, let $z_t^v$ denote the logit for token $v$ conditioned on context $y_{<t}$. We have the logit-level gradients for each token~\citep{zhu2026surprising,jia2026asymmetric}: 
\begin{equation}\label{eq:logit_gradient}
    -\frac{\partial\mathcal{L}_{\text{OPD}}}{\partial z_t^v}\propto \begin{cases}
        A_t\left(1-\pi_s(v\mid x;y_{<t})\right), & \text{if }v=y_t\text{ (sampled token)}\\
        -A_t(\pi_s(v\mid x;y_{<t})), & \text{if }v\neq y_t\text{ (unsampled token)}
    \end{cases}.
\end{equation}
We observe that the gradient vanishes in two regimes: when $|A_t|$ is small, and when $\pi_s(y_t\mid x,y_{<t})$ approaches one for the sampled token. Tokens in either regime have negligible impact on training.

\begin{figure}[t]
    \centering
    \includegraphics[width=1.0\linewidth]{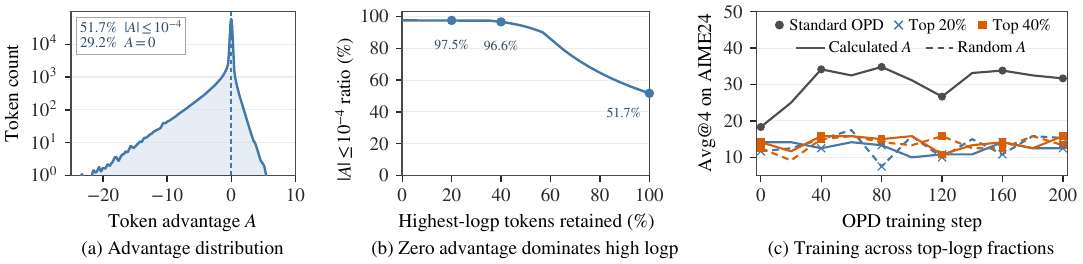}
    \caption{Qwen3-1.7B student with 4B-Instruct teacher. (a): Token-level advantage distribution during OPD, with most tokens receiving near-zero advantages. (b): Fraction of near-zero advantages among tokens within different top-percentile ranges of student logp. (c): Training only on top-logp tokens yields limited improvement, suggesting that these tokens provide weak learning signals.}\label{fig:top_logp_bad}
\end{figure}

\paragraph{Token Mass Concentrates at Near-Zero Advantages.}
We visualize the distribution of token-level advantages in Fig.~\ref{fig:top_logp_bad}(a). A substantial fraction of the tokens is concentrated near zero: $29.2\%$ of tokens have exactly zero advantage, and $51.7\%$ have advantage magnitude below $10^{-4}$.

\paragraph{Near-Zero Advantages Coincide with the Student's High-Logp Tokens.} 
Fig.~\ref{fig:top_logp_bad}(b) shows that tokens with near-zero advantages ($|A|<10^{-4}$) are concentrated among those assigned high log probability by the student. When the student is extremely confident in a token, the teacher, conditioned on the same student-generated prefix, often assigns similarly high probability to that token, resulting in only a negligible difference between their logp and, consequently, an advantage close to zero.

\paragraph{High-Logp Tokens Contribute Nothing to On-Policy Training.}
To test whether high-logp tokens provide any effective learning signal,  we conduct controlled on-policy training experiments. Specifically, we train on varying proportions of the student’s top-logp tokens using either the original OPD advantages (calculated A) or random advantages (random A) sampled evenly from $[-1,1]$. As shown in Fig.~\ref{fig:top_logp_bad}(c), restricting training to these tokens yields no noticeable improvement in the student’s AIME24 performance. Interestingly, the performance remains essentially unchanged even when the original advantages are replaced with random values. These results suggest that high-logp tokens provide no effective learning signal during on-policy training, largely regardless of the advantages assigned to them.

\subsection{Which Learning Signals Drive Policy Improvement?}\label{sec:3.2}
Revisiting the formulation of the OPD advantage in Eq.~\ref{eq:opd_loss}, the reverse KL is computed on student-sampled trajectories, which are inherently off-policy for the teacher. The teacher therefore assigns lower probabilities to many sampled tokens, resulting in predominantly negative advantages, consistent with the results in Fig.~\ref{fig:top_logp_bad}(a). 
A related observation appears in the RL setting as well, where \cite{zhu2026surprising} show that Negative Sample Reinforcement (NSR), which learns only from negative signals, can improve policy performance. This motivates us to ask whether the gains achieved by OPD are driven primarily by its large proportion of negative advantages, rather than by the specific supervision provided by the teacher.

We use \texttt{Qwen3-1.7B} as the student model and compare three advantage assignment schemes. The first follows standard OPD, using \texttt{Qwen3-4B-Instruct} as the teacher model to compute token-level advantages. The other two are teacher-free variants that assign a fixed advantage of $-0.5$ or $+0.2$ to each selected token. Since high-logp tokens provide little effective training gradient (Section~\ref{sec:3.1}), we restrict training to the 20\% of tokens with the lowest student logp.

\begin{figure}[t]
    \centering
    \includegraphics[width=1.0\linewidth]{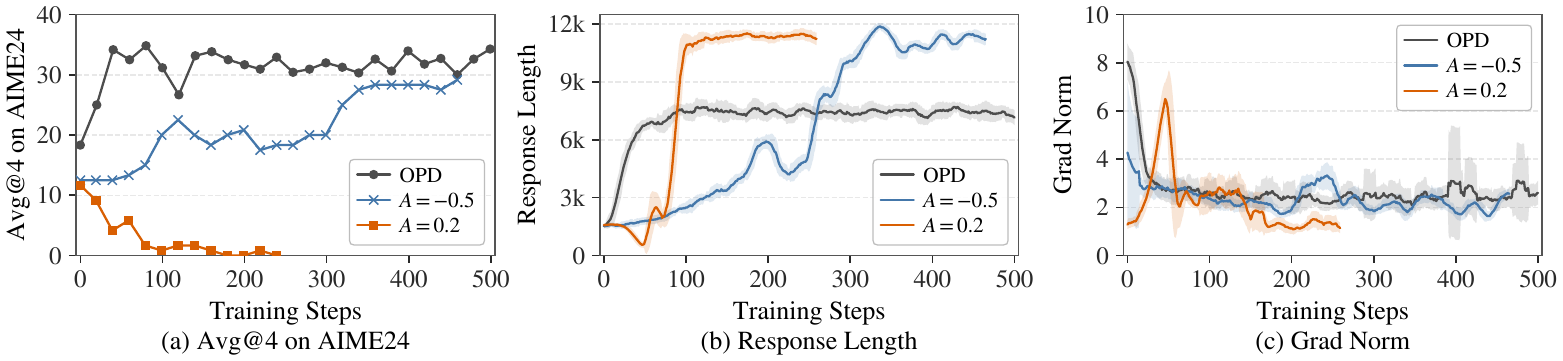}
    \caption{Training dynamics of standard OPD and on-policy training with fixed advantages.}
    \label{fig:fix_pos_neg}
\end{figure}

\paragraph{A Fixed Negative Advantage Alone Improves Students.}
We report the training dynamics of the different advantage-assignment schemes in Fig.~\ref{fig:fix_pos_neg}. First, OPD restricted to the 20\% of tokens with the lowest student log-probabilities achieves performance comparable to that of full-token OPD, further supporting the findings in Section~\ref{sec:3.1}. More notably, training with a fixed negative advantage yields a steady improvement in the student policy’s avg@4 performance on AIME24. Similar to standard OPD, the response length also increases gradually throughout training and eventually stabilizes at approximately 12K tokens. In contrast, training with a fixed positive advantage leads to policy collapse: during the first 40 training steps, the response length progressively decreases to nearly zero while the gradient norm explodes. After the collapse, the policy produces degenerate outputs consisting largely of seemingly random or garbled tokens.

\begin{question}{Takeaway}
    Our analysis suggests that OPD gains may arise not from distilling teacher knowledge, but from suppressing low-probability tokens sampled by the student itself, an operation that requires no teacher at all. This observation motivates the method introduced in Section~\ref{sec:4}.
\end{question}

\section{Methodology}\label{sec:4}
Section~\ref{sec:3} showed that OPD's improvements survive the removal of the teacher, of positive advantages, and of all but the lowest-logp tokens. The remaining question is: given that negative signal on low-logp tokens is what drives improvement, \emph{how much} negative signal should each such token receive? Section~\ref{sec:4.1} shows that token entropy determines the amount of signal; Section~\ref{sec:4.2} turns the findings into On-Policy Self-Adaptation (OPSA), a supervision-free training algorithm, and Section~\ref{sec:4.3} analyzes why it works.

\subsection{Entropy Determines the Amount of Negative Signals}\label{sec:4.1}
Section~\ref{sec:3.1} suggested that low-logp tokens are the ones that matter, but logp alone does not distinguish two very different situations. A token can have low log probability because the policy is uncertain, with probability mass spread across many plausible tokens, so that even head tokens have relatively low logp. Or the student distribution is sharply concentrated on only a few tokens, while the sampled token happens to fall in the tail of a confident distribution.
\cite{wang2026beyond,he2026hindsight} have shown that high-entropy tokens play an important role in reinforcement learning and may benefit from stronger learning signals. Motivated by these findings, we compare two entropy-aware schemes that dynamically adjust the magnitude of the negative advantage:
\begin{equation}\label{eq:dyn_adv}
    A_i^{\text{dyn}}=A_i^{\text{fix}}-\frac{1}{4}\delta\cdot r_i,\hspace{10pt}
    r_i=2\frac{H_i-H_{\text{min}}}{H_{\text{max}}-H_{\text{min}}}-1,
\end{equation}
where $H_{\min}$ and $H_{\max}$ are the minimum and maximum entropy over the lowest-$20\%$-logp positions within each response, so that $r_i\in[-1,1]$ measures a token's entropy relative to its own rollout. The parameter $\delta$ controls the relationship between the advantage magnitude $|A_i|$ and the token entropy $H_i$: $\delta=1$ assigns larger-magnitude negative advantages to higher-entropy tokens, $\delta=-1$ reverses this, and $\delta=0$ recovers the fixed-negative-advantage setting of Section~\ref{sec:3.2}.

\begin{figure}[t]
    \centering
    \includegraphics[width=1.0\linewidth]{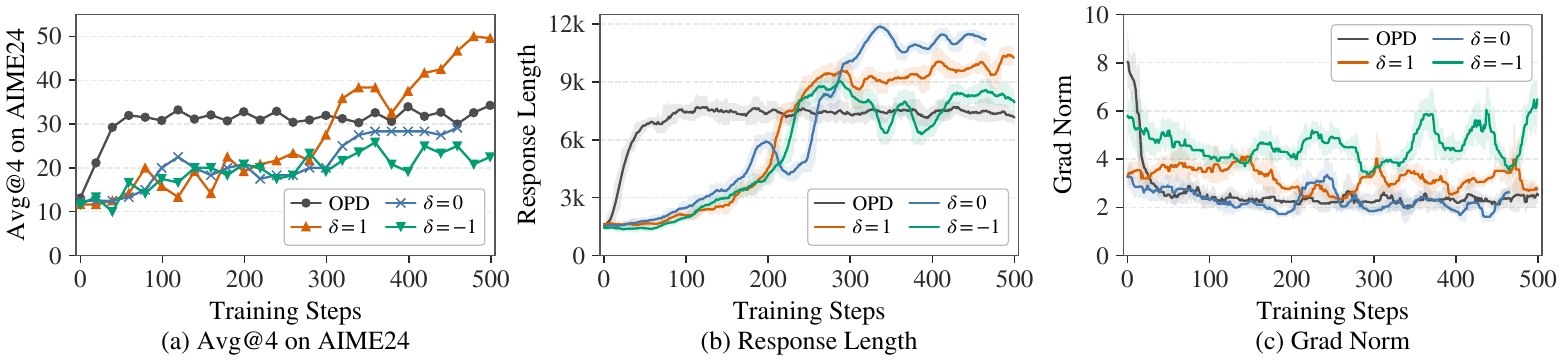}
    \caption{Training dynamics of on-policy training with dynamic negative advantages.}
    \label{fig:dynamic_adv}
\end{figure}
\paragraph{Adaptive Advantages with $\delta=1$ Outperform OPD.}
The results in Fig.~\ref{fig:dynamic_adv} show that, surprisingly, when the magnitude of the negative advantage is positively correlated with token entropy ($\delta=1$), the policy exhibits steady improvement in avg@4 on AIME24, ultimately reaching 50.0\% and substantially outperforming standard OPD at 35.13\%. Comparing the positively correlated ($\delta=1$) and negatively correlated ($\delta=-1$) variants further demonstrates that on-policy training benefits substantially from assigning larger-magnitude negative advantages to high-entropy tokens. In contrast, the negatively correlated variant becomes unstable between steps 350 and 450, maintains consistently higher gradient norms throughout training, and ultimately performs slightly worse than the fixed-negative-advantage baseline. These results highlight the importance of concentrating stronger learning signals on high-entropy tokens for effective and stable on-policy training.

\subsection{On-Policy Self-Adaptation}\label{sec:4.2}
These results give a complete on-policy training recipe, which we call \textbf{O}n-\textbf{P}olicy \textbf{S}elf-\textbf{A}daptation (\textbf{OPSA}). OPSA updates only the lowest-logp tokens, assigns them negative advantages, and scales the magnitude with entropy. It requires no external supervision of any kind, including teacher signals, ground-truth rewards, or reference answers.
Setting $A_i^{\text{fix}}=-\frac{3}{4}$ and $\delta=1$ in Eq.~(\ref{eq:dyn_adv}) and restricting the update to $\mathcal{S}_{\text{lowest}20}$, the training objective of OPSA is defined as follows:
\begin{equation}\label{eq:opsa_loss}
    \mathcal{L}_{\text{OPSA}}=-\mathbb{E}\left[\frac{1}{ \textcolor{myred}{\vert\mathcal{S}_{\text{lowest}20}\vert}}\sum_{\textcolor{myred}{i\in{\mathcal{S}}_{\text{lowest}20}}}\textcolor{myred}{A_i^{\text{dyn}}}\log\pi_\theta(y_i\vert x;y_{<i})\right], \hspace{4pt}\textcolor{myred}{A_i^{\text{dyn}}=-\frac{1}{2}-\frac{H_i-H_{\text{min}}}{2(H_{\text{max}}-H_{\text{min}})}}
\end{equation}

Since OPSA eliminates the need for a teacher model during training, it avoids additional forward passes through a teacher model to compute advantages. Instead, advantages are derived directly from the student policy’s token-level entropy, introducing negligible computational overhead and substantially reducing training time. Detailed efficiency results are reported in the Appendix~\ref{app:more_detail}.

\subsection{Why Does OPSA Work?}\label{sec:4.3}
\begin{figure}[t]
    \centering
    \includegraphics[width=1.0\linewidth]{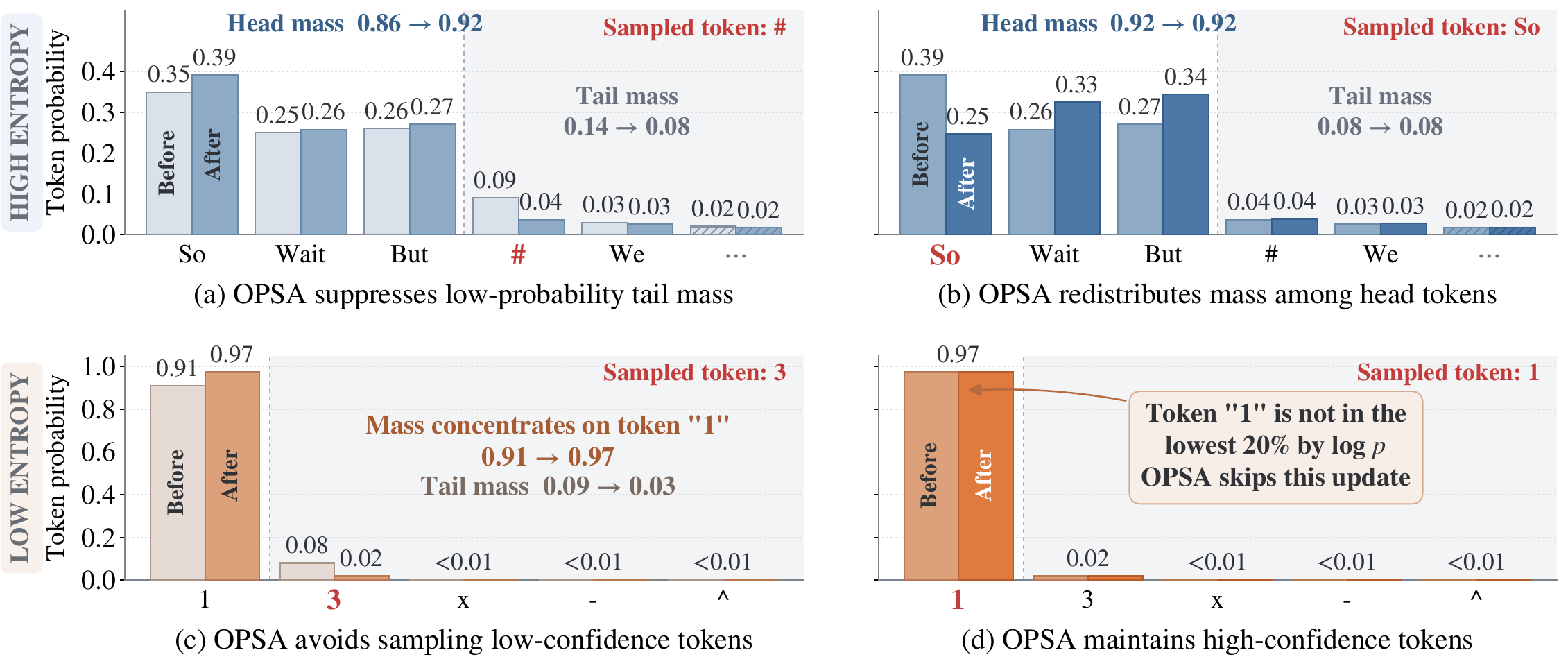}
    \caption{Examples of OPSA updates. (a) At high-entropy positions, OPSA suppresses sampled tail tokens and reallocates probability to head tokens. (b) At high-entropy positions, OPSA redistributes probability among head tokens, with little effect on the tail. (c) OPSA avoids sampling tail tokens at low-entropy positions. (d) OPSA preserves high-confidence predictions.}
    \label{fig:why_opsa_work}
\end{figure}

To further understand the mechanism of OPSA, we conduct a detailed analysis by categorizing tokens into four cases based on token entropy and sampled-token probability: (a) a tail token at a high-entropy position, (b) a head token at a high-entropy position, (c) a tail token at a low-entropy position, and (d) a head token at a low-entropy position. Based on the logit-level gradients in Eq.~\ref{eq:logit_gradient}, Fig.~\ref{fig:why_opsa_work} illustrates how the negative advantages in OPSA reshape the policy distribution in each case.

\paragraph{Tail Tokens: OPSA Suppresses Low-Probability Reasoning Branches.}
During on-policy training, stochastic sampling is typically used instead of greedy decoding to maintain rollout diversity. At relatively high sampling temperatures, however, extremely low-probability tail tokens may occasionally be sampled, causing the reasoning process to enter unlikely and potentially erroneous branches. Because OPSA adaptively assigns negative advantages to the 20\% of policy-generated tokens with the lowest logp, its policy update suppresses these sampled tail tokens and reallocates their probability mass to higher-probability head tokens, as illustrated in Fig.~\ref{fig:why_opsa_work}(a)(c). In this way, OPSA steers subsequent reasoning away from low-probability branches and toward more confident alternatives.

\paragraph{Head Tokens: OPSA Evenly Redistributes Mass Without Collapsing Diversity.}

Suppressing tail tokens makes sampling increasingly head-dominated, which raises the concern of diversity collapse. Fig.~\ref{fig:why_opsa_work}(b) shows why this does not occur. At a high-entropy position, the negative advantage assigned to a sampled head token redistributes probability mass among \emph{competing} head tokens rather than further sharpening the distribution onto a single prediction, while leaving the tail-token probabilities largely unaffected. This behavior preserves output diversity and the policy’s ability to explore alternative reasoning paths. Moreover, because OPSA updates only the 20\% of sampled tokens with the lowest student log-probabilities, high-confidence tokens at low-entropy positions are typically excluded from training. As shown in Fig.~\ref{fig:why_opsa_work}(d), OPSA therefore preserves confident predictions at low-entropy positions, where high precision is particularly important~\citep{zhang2026embarrassingly}.

\paragraph{Reshaped Distribution Enables Reflective Reasoning.}
Consistent with prior observations~\citep{dinglearning,zhao2025can}, we find that high token-level entropy is closely associated with self-reflective reasoning behavior. At such high-entropy positions, the model is more likely to generate reflective tokens such as ``wait'' and ``but'', which often signal reflection or self-correction and lead to higher-quality reasoning. The example in Fig.~\ref{fig:why_opsa_work} illustrates how OPSA increases the total probability mass of the head-token set at high-entropy positions while redistributing it more evenly within the set. This makes diverse reflective branches more likely to be explored. We provide a quantitative analysis of this phenomenon in Section~\ref{sec:5.3}.

\section{Experiments}
\subsection{Experimental Setup}
\paragraph{Models and Datasets.}
Our experiments cover models from the Qwen3~\citep{yang2025qwen3}, Qwen3.5~\citep{qwen35blog}, and Olmo3~\citep{olmo2025olmo} families. All models are trained on DAPO-17k~\citep{yu2026dapo} for mathematical reasoning tasks using only the questions, without access to labels and ground-truth answers. We evaluate the models on three in-domain mathematical reasoning benchmarks, AIME24, AIME25, and HMMT25, as well as the out-of-domain benchmark MBPP+~\citep{liu2023your} (Code), and GPQA-Diamond~\citep{rein2023gpqa} (Q\&A). We provide additional model details and code-task training results in Appendix~\ref{app:train_hyper} and~\ref{app:code_opsa_train}, respectively.

\begin{table}[t]
    \centering
    \small
    {
    \renewcommand{\arraystretch}{0.9}
    \begin{tabular}{lcccc}
    \toprule
    \textbf{Methods} & \textbf{Dense Signal}
    & \textbf{No Verifiable Reward}
    & \textbf{No External Teacher}
    & \textbf{No Hint} \\
    \midrule
    RLVR~\citep{shao2024deepseekmath}
    & \ding{55} & \ding{55} & \ding{51} & \ding{51} \\
    TTRL~\citep{zuo2026ttrl}
    & \ding{55} & \ding{51} & \ding{51} & \ding{51} \\
    OPD~\citep{lu2025onpolicydistillation}
    & \ding{51} & \ding{51} & \ding{55} & \ding{51} \\
    OPSD~\citep{zhao2026self}
    & \ding{51} & \ding{51} & \ding{51} & \ding{55} \\
    \rowcolor{cite!20}
    \textbf{OPSA (Ours)}
    & \ding{51} & \ding{51} & \ding{51} & \ding{51} \\
    \bottomrule
    \end{tabular}
    }
    \caption{Comparison of supervision signals in different on-policy training methods.}
    \label{tab:comparison_method}
\end{table}

\paragraph{Baselines.}
We compare OPSA with representative on-policy training methods, including reinforcement learning with verifiable rewards (RLVR), such as GRPO~\citep{shao2024deepseekmath}; Test-Time Reinforcement Learning (TTRL)~\citep{zuo2026ttrl}, which does not require external supervision; standard On-Policy Distillation (OPD)~\citep{lu2025onpolicydistillation}; and On-Policy Self-Distillation (OPSD)~\citep{zhao2026self}, which replaces an external teacher with hints corresponding to questions. Table~\ref{tab:comparison_method} summarizes the key properties of these methods. OPSA combines their desirable characteristics: it provides dense token-level learning signals while requiring neither verifiable rewards, external teacher models, nor auxiliary hints.

\paragraph{Implementation Details.}
We implement OPSA using the slime~\citep{slime_github} framework and optimize the objective in Eq.~\ref{eq:opsa_loss}. The loss is computed only over the 20\% of policy-sampled tokens with the lowest log-probabilities. All experiments are conducted on 8 NVIDIA H100 or H200 GPUs.

\subsection{Main Results}

\begin{table}[t]
    \centering
    \small
    \begin{tabular}{lc@{\hskip 5pt}cc@{\hskip 5pt}cc@{\hskip 5pt}cc@{\hskip 5pt}c}
    \toprule
    \multirow{3}{*}{\textbf{Methods}} & \multicolumn{6}{c}{\textbf{Math (I.D.)}} & \multicolumn{2}{c}{\textbf{O.O.D.}}\\
    \cmidrule(lr){2-7} \cmidrule(lr){8-9}
    & \multicolumn{2}{c}{\textbf{AIME24}}
    & \multicolumn{2}{c}{\textbf{AIME25}}
    & \multicolumn{2}{c}{\textbf{HMMT25}}
    & {\textbf{MBPP+}} & \textbf{GPQA$_{D}$}\\
    & avg@32 & pass@32
    & avg@32 & pass@32
    & avg@32 & pass@32
    & \multicolumn{2}{c}{avg@32} \\
    \midrule
    Qwen3-1.7B & 13.44 & 40.00 & 9.69 & 30.00 & 5.73 & 23.33 & 58.24 & 27.92 \\
    \rowcolor{cite!20}w/ OPSA & 48.85 & 80.00 & 35.31 & 66.67 & 23.33 & 50.00 & 59.44 & 32.40 \\ 
    \rowcolor{cite!20}$\Delta$ & \textcolor{myred}{+35.41} & \textcolor{myred}{+40.00} & \textcolor{myred}{+25.62}  & \textcolor{myred}{+36.67} & \textcolor{myred}{+17.60} & \textcolor{myred}{+26.67} & \textcolor{myred}{+1.20} & \textcolor{myred}{+4.48}\\
    & $\uparrow$ 263.5\% & $\uparrow$ 100.0\% & $\uparrow$ 264.4\% & $\uparrow$ 122.2\% & $\uparrow$ 307.2\% & $\uparrow$ 114.3\% & $\uparrow$ 2.1\% & $\uparrow$ 16.0\% \\
    \midrule
    Qwen3-4B & 23.33 & 56.67 & 20.52 & 56.67 & 13.13 & 33.33 & 66.93 & 38.46 \\ 
    \rowcolor{cite!20}w/ OPSA & 62.08 & 83.33 & 58.44 & 83.33 & 37.40 & 60.00 & 68.35 & 41.29 \\
    \rowcolor{cite!20}$\Delta$ & \textcolor{myred}{+38.75} & \textcolor{myred}{+26.66} & \textcolor{myred}{+37.92} & \textcolor{myred}{+26.66} & \textcolor{myred}{+24.27} & \textcolor{myred}{+26.67} & \textcolor{myred}{+1.42} & \textcolor{myred}{+2.83} \\
    & $\uparrow$ 166.1\% & $\uparrow$ 47.0\% & $\uparrow$ 184.8\% & $\uparrow$ 47.0\% & $\uparrow$ 184.8\% & $\uparrow$ 80.0\% & $\uparrow$ 2.1\% & $\uparrow$ 7.4\% \\
    \midrule
    Qwen3.5-9B & 76.35 & 93.33 & 56.04 & 93.33 & 44.48 & 86.67 & 77.33 & 70.53 \\
    \rowcolor{cite!20}w/ OPSA & 87.81 & 96.67 & 76.98 & 96.67 & 67.40 & 93.33 & 79.27 & 73.70 \\
    \rowcolor{cite!20}$\Delta$ & \textcolor{myred}{+11.46} & \textcolor{myred}{+3.34} & \textcolor{myred}{+20.94} & \textcolor{myred}{+3.34} & \textcolor{myred}{+22.92} & \textcolor{myred}{+6.66} & \textcolor{myred}{+1.94} & \textcolor{myred}{+3.17} \\
    & $\uparrow$ 15.0\% & $\uparrow$ 3.6\% & $\uparrow$ 37.4\% & $\uparrow$ 3.6\% & $\uparrow$ 51.5\% & $\uparrow$ 7.7\% & $\uparrow$ 2.5\% & $\uparrow$ 4.5\% \\
    \midrule
    Olmo3-7B$_\text{IT}$ & 54.48 & 83.33 & 38.75 & 76.67 & 24.90 & 56.67 & 57.79 & 34.12 \\
    \rowcolor{cite!20} w/ OPSA & 61.88 & 93.33 & 47.92 & 83.33 & 29.69 & 63.33 & 59.61 & 35.64 \\
    \rowcolor{cite!20} $\Delta$ & \textcolor{myred}{+7.40} & \textcolor{myred}{+10.00} & \textcolor{myred}{+9.17} & \textcolor{myred}{+6.67} & \textcolor{myred}{+4.79} & \textcolor{myred}{+6.66} & \textcolor{myred}{+1.82} & \textcolor{myred}{+1.52} \\
    & $\uparrow$ 13.6\% & $\uparrow$ 12.0\% & $\uparrow$ 23.7\% & $\uparrow$ 8.7\% & $\uparrow$ 19.2\% & $\uparrow$ 11.8\% & $\uparrow$ 3.1\% & $\uparrow$ 4.4\% \\
    \midrule
    Olmo3-7B$_\text{TK}$ & 71.88 & 93.33 & 61.77 & 90.00 & 41.98 & 73.33 & 64.55 & 38.86 \\
    \rowcolor{cite!20} w/ OPSA & 77.29 & 93.33 & 67.08 & 93.33 & 47.50 & 83.33 & 66.33 & 40.93 \\
    \rowcolor{cite!20} $\Delta$ & \textcolor{myred}{+5.41} & \textcolor{myred}{+0.00} & \textcolor{myred}{+5.31} & \textcolor{myred}{+3.33} & \textcolor{myred}{+5.52} & \textcolor{myred}{+10.00} & \textcolor{myred}{+1.78} & \textcolor{myred}{+2.07} \\
    & $\uparrow$ 7.5\% & $\uparrow$ 0.0\% & $\uparrow$ 8.6\% & $\uparrow$ 3.7\% & $\uparrow$ 13.1\% & $\uparrow$ 13.6\% & $\uparrow$ 2.8\% & $\uparrow$ 5.3\% \\
    \bottomrule
    \end{tabular}
    \caption{Performance of OPSA across models on in-domain math and out-of-domain code and Q\&A tasks. Qwen models use non-thinking mode; Olmo ${\text{IT}}$ and ${\text{TK}}$ denote Instruct and Think variants.}
    \label{tab:opsa_main}
\end{table}

\paragraph{OPSA Consistently Improves Different Models.}
Table~\ref{tab:opsa_main} demonstrates that OPSA consistently improves performance across model families, scales, and reasoning modes. For the Qwen3 series, OPSA more than doubles Avg@32 on every mathematical reasoning benchmark. In particular, for \texttt{Qwen3-1.7B}, it yields relative improvements of 263\% on AIME24 and AIME25 and 307\% on HMMT25. The gains remain substantial on the stronger \texttt{Qwen3.5-9B}, where OPSA further improves Avg@32 by 11.46 points on AIME24 and 22.92 points on HMMT25. We observe the same consistent trend on two \texttt{Olmo3-7B} variants, despite their substantially different baseline performance, with OPSA improving Avg@32 and Pass@32 across nearly all benchmarks.
OPSA also consistently improves Pass@32. Across the Qwen3 series, relative gains range from 47\% to 122\%. Even for stronger models with near-saturated Pass@32, such as \texttt{Qwen3.5-9B} and \texttt{Olmo3-7B}$_{\text{TK}}$, OPSA maintains or further improves Pass@32, including gains of 6.66 points on HMMT25 for \texttt{Qwen3.5-9B} and 10.00 points for \texttt{Olmo3-7B}$_{\text{TK}}$. These results demonstrate that OPSA remains effective across models with substantially different initial capabilities.

\begin{table}[t]
    \centering
    \small
    \begin{tabular}{l@{\hskip 7pt}c@{\hskip 5pt}cc@{\hskip 5pt}cc@{\hskip 5pt}cc@{\hskip 5pt}c}
    \toprule
    \multirow{2}{*}{\textbf{Methods}} & \multicolumn{2}{c}{\textbf{AIME24}}
    & \multicolumn{2}{c}{\textbf{AIME25}}
    & \multicolumn{2}{c}{\textbf{HMMT25}}
    & \multicolumn{2}{c}{\textbf{Average}}\\
    & avg@32 & pass@32
    & avg@32 & pass@32
    & avg@32 & pass@32
    & avg@32 & pass@32\\
    \midrule
    Qwen3-1.7B & 13.44 & 40.00 & 9.69 & 30.00 & 5.73 & 23.33 & 9.62 & 31.11 \\
    + GRPO & 33.96 & 70.00 & 25.31 & 50.00 & 15.10 & 43.33 & 24.79 & 54.44 \\
    + TTRL & 19.90 & 30.00 & 9.79 & 30.00 & 5.73 & 23.33 & 11.81 & 27.78 \\
    + OPD & 32.08 & 73.33 & 20.52 & 50.00 & 13.85 & 40.00 & 22.15 & 54.44 \\
    + OPSD & 33.33 & 73.33 & 22.50 & 53.33 & 14.90 & 43.33 & 23.58 & 56.67 \\
    \rowcolor{cite!20}+ OPSA & 48.85 & 80.00 & 35.31 & 66.67 & 23.33 & 50.00 & 35.83 & 65.56 \\ 
    \rowcolor{cite!20}$\Delta$=OPSA-RL$_{best}$ & \textcolor{myred}{+14.89} & \textcolor{myred}{+6.67} & \textcolor{myred}{+10.00} & \textcolor{myred}{+13.34} & \textcolor{myred}{+8.23} & \textcolor{myred}{+6.67} & \textcolor{myred}{+11.04} & \textcolor{myred}{+8.89} \\ 
    \midrule
    Qwen3-1.7B$_{\text{Thinking}}$ & 46.56 & 80.00 & 36.25 & 73.33 & 22.92 & 60.00 & 35.24 & 71.11 \\
    \rowcolor{cite!20}+ OPSA & 52.50 & 83.33 & 42.79 & 73.33 & 26.85 & 60.00 & 40.71 & 72.22 \\
    \bottomrule
    \end{tabular}
    \caption{Comparison of different on-policy RL methods on \texttt{Qwen3-1.7B}.}
    \label{tab:compare_basline}
\end{table}

\paragraph{OPSA Outperforms Baselines Using Zero External Supervision.}
We compare OPSA with several representative on-policy RL baselines. For clarity, TTRL performs test-time training directly on AIME24, and we evaluate its best checkpoint; OPD uses \texttt{Qwen3-4B-Instruct} as the teacher model. 
As shown in Table~\ref{tab:compare_basline}, OPSA consistently outperforms all baselines in both Avg@32 and Pass@32 while requiring no external supervision. Averaged across the three benchmarks, OPSA surpasses the best baseline by 11.04 points in Avg@32 and 8.89 points in Pass@32.
Although TTRL is also free from external supervision, its self-consistency-based training tends to sharpen the policy distribution around a local optimum, substantially degrading Pass@$k$. OPSD removes the external teacher by conditioning the student policy on additional hints to construct a teacher distribution. However, we find that it yields meaningful gains only when thinking mode is disabled for the student but enabled for the teacher, thereby creating a substantial distributional mismatch at the first token position. OPD and GRPO, two widely adopted post-training methods, achieve comparable overall performance. As shown in Table~\ref{tab:overhead} in Appendix~\ref{app:more_detail}, OPD, OPSD, and OPSA require fewer rollout generations during training and are therefore substantially more efficient than GRPO.

\paragraph{Generalization of OPSA.}
To evaluate its out-of-domain generalization, we further test the models on the code-generation benchmark MBPP+ and general Q\&A benchmark GPQA-Diamond. As shown in Table~\ref{tab:opsa_main}, OPSA consistently improves both tasks across different models, demonstrating its ability to generalize beyond the training domain.
Additionally, we disable thinking mode (\texttt{enable\_thinking=false}) during the OPSA training on Qwen models. Table~\ref{tab:compare_basline} also reports performance when thinking mode is enabled at inference time. Interestingly, with thinking mode disabled, the OPSA model achieves performance comparable to that of the base model with thinking mode enabled, suggesting that OPSA shifts the policy toward longer-form reasoning behaviors. Enabling thinking mode for the OPSA model yields further improvements, demonstrating that the gains from OPSA remain complementary to the model’s native thinking capability.

\subsection{More Analysis and Ablation}

\subsubsection{OPSA Elicits Reflective Long-Form Reasoning}\label{sec:5.2}
\begin{figure}[h]
    \centering
    \includegraphics[width=1.0\linewidth]{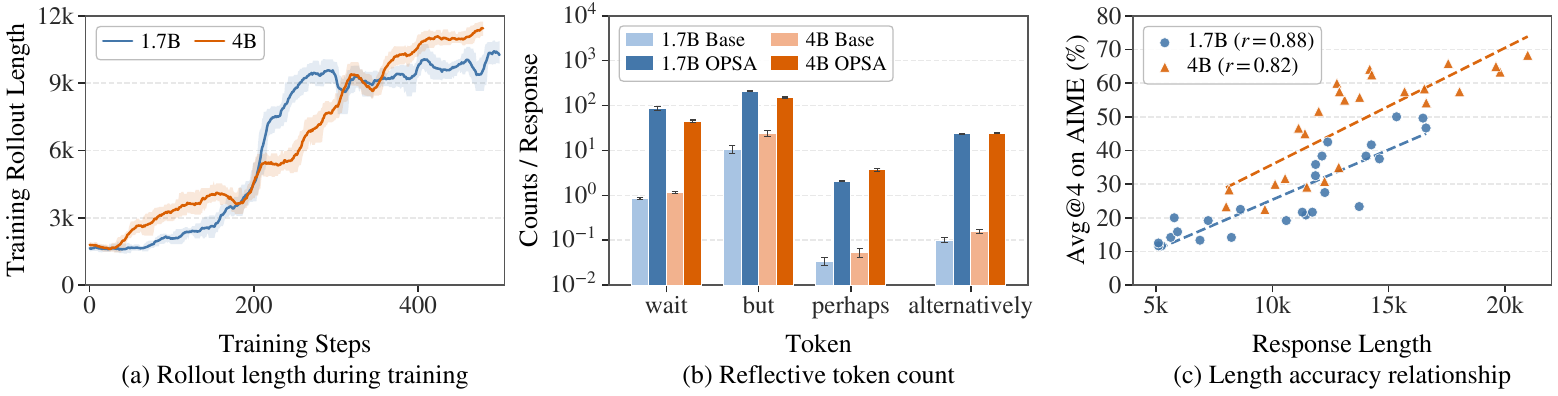}
    \caption{OPSA elicits long-form reasoning by generating more reflective tokens compared to base models. Moreover, AIME24 Avg@4 performance increases positively with response length.}
    \label{fig:reason_len}
\end{figure}

Section~\ref{sec:4.3} shows that, at high-entropy positions, OPSA redistributes probability mass among head tokens rather than concentrating it on a single prediction. Such positions often correspond to ``fork'' tokens in the chain of thought~\citep{wang2026beyond,zhang2026embarrassingly}, where different tokens may initiate distinct reasoning branches. Fig.~\ref{fig:reason_len}(a) shows that the response length increases steadily throughout OPSA training.
To better understand this change, we compare the frequency of reflective tokens in reasoning trajectories before and after training. As shown in Fig.~\ref{fig:reason_len}(b), the OPSA-trained model produces substantially more reflective expressions, indicating more frequent reflection and self-correction during reasoning. We further observe a clear positive correlation between response length and answer accuracy. Together, these findings suggest that OPSA promotes exploration at high-entropy forks, leading the model toward longer and potentially more productive reasoning branches characterized by increased reflection and self-correction.
\paragraph{Performance Drops When Fork Tokens Are Masked.}
\begin{wrapfigure}{r}{0.4\textwidth}
    \centering
    \vspace{-1.5em}
\includegraphics[width=\linewidth]{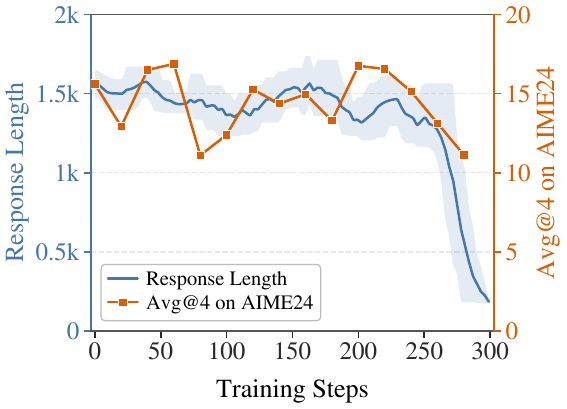}\vspace{-5pt}
    \caption{Training dynamics of OPSA when masking fork tokens.}
    \label{fig:mask_fork}
\end{wrapfigure}
To validate whether OPSA’s gains primarily arise from probability redistribution at fork tokens in the chain of thought, we conduct an ablation study by masking fork positions whose head-token sets contain reflective words, thereby excluding these positions from training. 
As shown in Fig.~\ref{fig:mask_fork}, masking these fork tokens largely eliminates the increases in both response length and accuracy observed under OPSA. Moreover, the response length collapses at approximately 300 training steps. 
These results demonstrate that applying OPSA at fork positions suppresses low-probability tail tokens while redistributing probability mass among competing head tokens. This encourages the policy to explore longer and more reflective reasoning branches, which appears to be a major source of its performance improvement. More details are given in Appendix~\ref{app:more_detail}.

\subsubsection{OPSA Doesn't Hurt Models' Diversity}\label{sec:5.3}
\begin{figure}[h]
    \centering
    \includegraphics[width=1.0\linewidth]{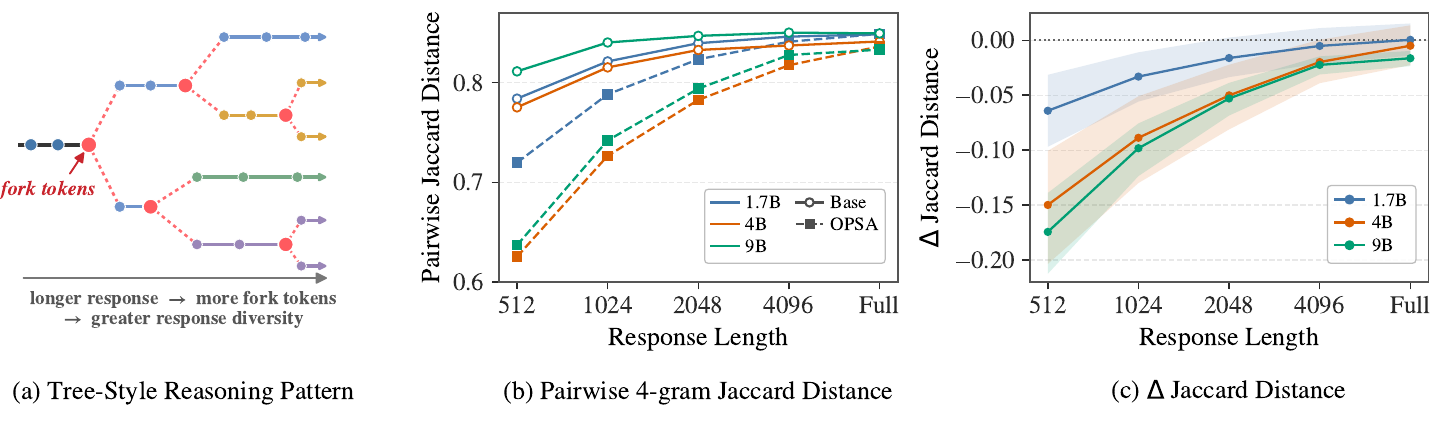}
    \caption{Diversity analysis of OPSA. (a) As reasoning length increases, more opportunities for branching arise. At each fork, OPSA distributes probability more evenly across alternative tokens, allowing multiple sampling to explore different reasoning paths and thereby increasing response diversity.  (b–c) Jaccard-distance comparisons on AIME24 before and after OPSA training, using 32 sampled responses per problem.}
    \label{fig:diversity}
\end{figure}
Although the analysis in Section~\ref{sec:4.3} shows that OPSA sharpens the overall token distribution, particularly at low-entropy positions, it redistributes probability mass among competing head tokens at high-entropy fork positions. This mechanism allows OPSA to preserve response diversity in long-form reasoning. We quantify diversity using Jaccard distance (JD)~\citep{broder1997resemblance,wang2018sentigan}, where lower values indicate greater similarity between responses. Fig.~\ref{fig:diversity} reports JD for the base and OPSA-trained models across different response lengths. As the number of generated tokens increases, the JD gap between the two models gradually narrows and eventually approaches zero, indicating that OPSA preserves a level of long-form response diversity comparable to that of the base model. This result suggests that OPSA continues to explore alternative reasoning branches at successive fork positions, producing the branching, tree-like reasoning patterns illustrated in Fig.~\ref{fig:diversity}(a) without causing diversity collapse. The Pass@32 results in Table~\ref{tab:opsa_main} provide further evidence that OPSA-induced distribution sharpening neither restricts the policy’s exploration space nor degrades its pass@k performance.

\begin{wrapfigure}{r}{0.43\textwidth}
    \centering
    \vspace{-1em}
\includegraphics[width=\linewidth]{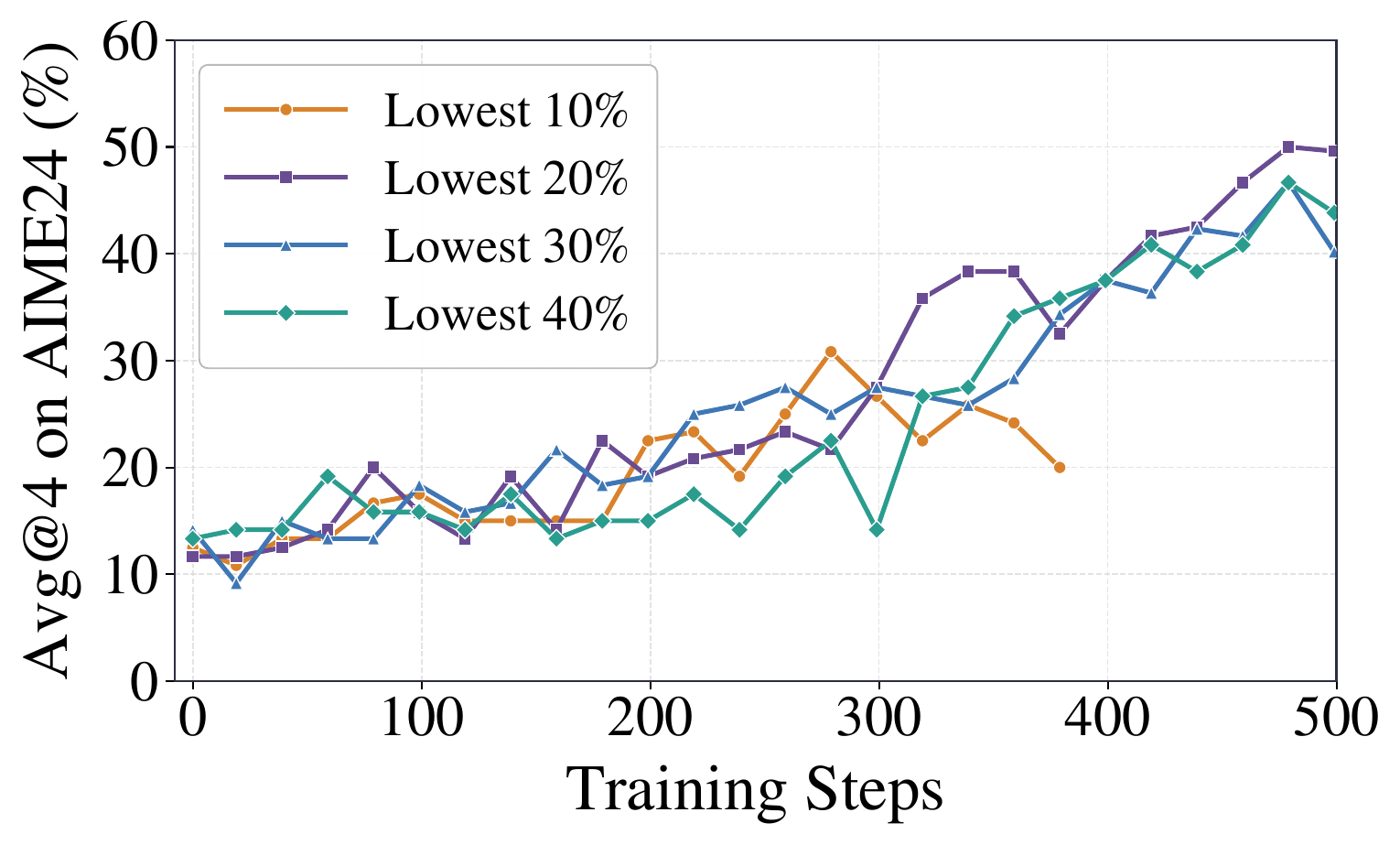}\vspace{-5pt}
    \caption{Ablation study of OPSA with different training token ratios.}
    \label{fig:ablation_low}
\end{wrapfigure}

\subsubsection{Ablation Study of The Fraction of Trained Tokens}
We have examined the advantage-assignment scheme of OPSA in Section~\ref{sec:3.2}, where negative advantages positively correlated with token entropy yield the strongest improvement. Here, we ablate the proportion of low-log-probability tokens used for training. Fig.~\ref{fig:ablation_low} reports the Avg@4 training curves when OPSA is applied to the lowest 10\%, 20\% (ours), 30\%, and 40\% of tokens ranked by their on-policy log-probabilities.

Training on only the bottom 10\% performs substantially worse than the other settings. We find that these tokens consist almost entirely of low-probability tail tokens outside the top-1 prediction. Applying negative advantages exclusively to them over-sharpens the policy distribution and causes a pronounced entropy decrease, thereby limiting further improvement. In contrast, the 20\%, 30\%, and 40\% settings all increase Avg@4 to above 45. These results indicate that OPSA is not highly sensitive to the exact token-selection ratio and that training on only the bottom 20\% of sampled tokens is sufficient to achieve substantial gains.

\section{Related Works}

\paragraph{Supervision Signals in On-Policy Distillation.}
OPD optimizes a student policy on its self-generated trajectories. GKD~\citep{agarwal2024policy} combines student-generated and supervised trajectories under flexible divergence objectives, while MiniLLM~\citep{gu2024minillm} adopts reverse-KL distillation to reduce exposure bias. More recently, \citet{lu2025onpolicydistillation} improve OPD efficiency for reasoning tasks using the K1 estimator. Another line of work~\citep{zhao2026self,shenfeld2026self,hubotter2026reinforcement} replaces the external teacher with the same policy conditioned on additional hints, yielding on-policy self-distillation (OPSD). However, OPSD retains the OPD training paradigm, leaving unresolved whether the constructed teacher can provide reliable supervision on student-generated trajectories~\citep{kaur2026rethinking}. Recent studies aim to avoid potential training noise in OPD through token selection~\citep{xu2026tip,fu2026revisiting} or fine-grained credit assignment~\citep{xing2026trust,xie2026position,hou2026uni}, but still rely on external supervision. In contrast, we investigate the source of OPD’s gains and propose OPSA, which enables self-improvement without external supervision.

\paragraph{Label-Free Policy Improvement.}
Label-free training enables a policy to improve using signals derived from the model itself, without external supervision. Self-rewarding methods~\citep{yuan2024self,ding2026sherlock} use an LLM-as-a-Judge~\citep{zheng2023judging} to score the model’s own responses and construct reward signals. TTRL~\citep{zuo2026ttrl} and EMPO~\citep{zhang2026right} instead infer pseudo-gold answers from GRPO rollouts through majority voting or clustering, and then use them for reward assignment. Intuitor~\citep{zhao2025learning} adopts self-certainty, using trajectory-level entropy as a reward to encourage the policy toward more confident outputs. 
However, these methods depend heavily on the policy’s initial capability. If the correct answer does not lie near a mode of the policy distribution, coarse trajectory-level rewards may reinforce an incorrect mode, causing the policy to become increasingly overconfident and ultimately leading to a collapse in Pass@$k$. In contrast, OPSA does not force the policy to converge toward any self-generated reward signal. Instead, it provides finer-grained token-level learning signals through entropy-adaptive negative advantages, simultaneously improving precision at low-entropy positions and exploration at high-entropy fork tokens, thereby enhancing the policy’s overall reasoning capability.

\section{Conclusion}

The central contribution of this work is a systematic investigation into the source of performance gains in OPD. We find that much of OPD’s improvement can be reproduced without teacher supervision by assigning fixed negative advantages to low-probability tokens, while scaling these advantages according to token-level entropy further enhances performance. Building on these findings, we introduce \textbf{O}n-\textbf{P}olicy \textbf{S}elf-\textbf{A}daptation (\textbf{OPSA}), an external-supervision-free framework that enables policy self-improvement through entropy-adaptive negative advantages.

Our analysis shows that OPSA suppresses low-probability tail tokens while redistributing probability mass among competing head tokens. This adaptively sharpens the overall token distribution, improving prediction confidence at low-entropy positions while preserving exploration at high-entropy reasoning forks. As a result, OPSA encourages longer and more reflective reasoning trajectories, leading to substantial improvements in both Avg@32 and Pass@32.

Notably, OPSA outperforms OPD without teacher supervision while exhibiting similar training dynamics, including response-length growth. This observation calls for a reexamination of the mechanism underlying OPD with the K1 estimator: its gains may arise not only from making the student imitate a stronger teacher, but also from reshaping the student policy's own probability distribution. More broadly, our results show that the policy's internal token-level uncertainty can itself provide an effective fine-grained learning signal for on-policy self-improvement, without rewards, reference answers, or teacher-provided supervision.

\bibliography{main}
\bibliographystyle{main}

\newpage

\appendix

\section{Limitations and Future Directions}
Due to limited computational resources, our experiments focus on relatively small models of up to 9B parameters. It therefore remains unclear whether OPSA scales effectively to larger models or mixture-of-experts architectures. Moreover, our analysis suggests that OPSA primarily operates by redistributing the policy’s existing probability mass. Its benefits may therefore be limited for heavily post-trained models whose output distributions are already overly sharp and exhibit very low entropy. Although OPSA requires no external annotations and can serve as a plug-and-play post-training method, it may not substantially expand the policy’s underlying exploration frontier, as reflected by the relatively modest improvements in thinking-mode Pass@$k$. A promising direction is to combine OPSA’s fine-grained negative-advantage assignment with other reinforcement learning methods to further improve policy exploration.
In addition, OPSA outperforms OPD without teacher supervision while exhibiting similar training dynamics, including response-length growth. This finding calls for a reexamination of the mechanism underlying OPD with the K1 estimator and motivates further theoretical analysis to identify the true source of OPD's performance gains.
\section{Experimental Details}

\subsection{Training Configuration}\label{app:train_hyper}

\begin{table}[h]
    \small
    \centering
    \begin{tabular}{ll}
    \toprule
        \textbf{Hyperparameter} & \textbf{Value} \\
        \midrule
        Framework & \texttt{slime} (\texttt{0.2.4})\\
        Training engine & Megatron (\texttt{0.16.0rc0}) \\
        Rollout engine & SGLang (\texttt{0.5.14})\\
        GPUs & 8xH100 GPUs \\
        Learning rate & 1$\times$10$^{-6}$\\
        Rollout batch size & 64 \\
        n samples per prompt & 1 \\
        Enable Thinking & \texttt{False} \\
        Global batch size & 64 \\
        Training decoding & Temperature 1.0, top-k -1, top-p 1.0, max response length 12000\\
        Evaluation decoding & Temperature 0.7, top-k 20, top-p 0.8, max response length 32768\\
        Save \& Evaluation frequency & Save every 20 steps, validate every 20 steps\\
        Checkpoint selection metric & Avg@4 on validation set \\
        \bottomrule
        
    \end{tabular}
    \caption{Training hyperparameters for OPSA.}
    \label{tab:train_configuration}
\end{table}

\begin{table}[h]
    \small
    \centering
    \begin{tabular}{lcc}
    \toprule
        \textbf{Models} & \textbf{Default Thinking Mode} & \textbf{Training Thinking Mode} \\
        \midrule
        \texttt{Qwen3-1.7B} & Hybrid & Non-Thinking\\
        \texttt{Qwen3-4B} & Hybrid & Non-Thinking\\
        \texttt{Qwen3.5-9B} & Hybrid & Non-Thinking\\
        \texttt{Olmo3-7B-Instruct} & Non-Thinking & Non-Thinking\\
        \texttt{Olmo3-7B-Think} & Thinking & Thinking\\
        \bottomrule
        
    \end{tabular}
    \caption{Detailed training setting for different models.}
    \label{tab:model_config}
\end{table}
    
We provide the detailed OPSA training configuration in Table~\ref{tab:train_configuration}. For rollout during training, we adopt the default decoding parameters from the \texttt{slime} repository. For evaluation, we use the decoding parameters recommended in the Qwen3 model card. Following prior work~\citep{zhao2026self,hou2026uni}, we select the best checkpoint for each method based on Avg@4 performance.
To avoid ambiguity arising from different reasoning modes across models, Table~\ref{tab:model_config} summarizes each model’s default reasoning mode and the mode used for rollout generation during training.

\subsection{Evaluation Configuration}

\begin{table}[h]
    \small
    \centering
    \begin{tabular}{ll}
    \toprule
        \textbf{Hyperparameter} & \textbf{Value} \\
        \midrule
        Rollout engine & SGLang (\texttt{0.5.14})\\
        Temperature & 0.7 \\
        Top-k & 20 \\
        Top-p & 0.8 \\
        Max new tokens & 32768\\
        Enable thinking & \texttt{False} \\
        Samples per prompt & 32 \\
        Metric & Avg@32 \& Pass@32 \\
        \bottomrule
        
    \end{tabular}
    \caption{Evaluation hyperparameters for OPSA.}
    \label{tab:eval_configuration}
\end{table}

We provide the detailed evaluation configuration in Table~\ref{tab:eval_configuration}. We use \texttt{SGLang} for inference, generate 32 responses per prompt with thinking mode disabled, and report both Avg@32 and Pass@32.

\subsection{More Details}\label{app:more_detail}

\paragraph{Training and Inference Overhead of OPSA.}

\begin{table}[h]
    \small
    \centering
    \begin{tabular}{lcccc}
    \toprule
         & \textbf{Training} & \multicolumn{3}{c}{\textbf{Inference}} \\
        \textbf{Methods} & \textbf{Step Time (s)} & \textbf{Token Counts} & \textbf{Time (s)} & \textbf{Avg@32} \\
        \midrule
        Qwen3-1.7B & - & 4457 & 1.78 & 13.44 \\
        + GRPO & 186.2 & 19108 & 5.27 & 33.96 \\
        + OPD & 61.2 & 15286 & 5.73 & 32.08 \\
        + OPSA & 46.3 & 23205 & 6.58 & 48.85 \\
        \midrule
        Qwen3-4B & - & 8015 & 3.31 & 23.33 \\
        + OPSA & 68.6 & 20972 & 6.97 & 62.08 \\
        \midrule
        Qwen3.5-9B & - & 6847 & 5.29 & 76.35 \\
        + OPSA & 214.7 & 9695 & 6.31 & 87.81 \\
        \bottomrule
        
    \end{tabular}
    \caption{Training and inference overhead of OPSA compared to baselines.}
    \label{tab:overhead}
\end{table}

We compare the training and inference overhead of different methods in Table~\ref{tab:overhead}. OPSA and OPD train faster than GRPO because they do not require extensive rollout sampling to construct response groups. OPSA is even more efficient than OPD, as it neither deploys an additional teacher model nor performs forward passes through a larger and slower teacher. This allows more GPUs to be allocated to policy rollout and optimization. At inference time, OPSA steers the policy toward more reflective reasoning branches, producing longer responses while achieving higher accuracy.

\paragraph{Reflective Word Set for Experiments in Fig.~\ref{fig:mask_fork}.}
We list the detailed reflective word set used in experiments in Fig.~\ref{fig:mask_fork} as follows:

\begin{question}{Reflective Word Set}
\texttt{"wait", "however", "but", "alternatively", "hmm", "perhaps", "check", "might", "actually"}
\end{question}
Following the standard OPSA training procedure, we first identify the 20\% of token positions with the lowest student log-probabilities. At each selected position, we inspect the student’s top-5 candidate tokens. If any candidate belongs to the predefined reflective word set, we classify the position as a reflective fork and exclude it from the training objective.

\paragraph{Jaccard Distance in Section~\ref{sec:5.3}.}
We measure response diversity using the pairwise Jaccard distance between
sets of token-level 4-grams. For each response, we retain its first \(L\)
tokens, where \(L \in \{512, 1024, 2048, 4096, \mathrm{Full}\}\), and denote
the resulting set of unique 4-grams by
\(\mathcal{G}_4(r^{(\leq L)})\). The distance between two responses is
defined as
\begin{equation}
d_{\mathrm{J}}(r_i,r_j;L)
=
1-
\frac{
\left|
\mathcal{G}_4(r_i^{(\leq L)})
\cap
\mathcal{G}_4(r_j^{(\leq L)})
\right|
}{
\left|
\mathcal{G}_4(r_i^{(\leq L)})
\cup
\mathcal{G}_4(r_j^{(\leq L)})
\right|
}.
\end{equation}
For each AIME24 problem, we average the distance over all response pairs
among the \(R=32\) sampled responses, and then report the macro-average
over all \(P=30\) problems:
\begin{equation}
D_{\mathrm{J}}(L)
=
\frac{1}{P}
\sum_{p=1}^{P}
\frac{2}{R(R-1)}
\sum_{1 \leq i < j \leq R}
d_{\mathrm{J}}(r_{p,i},r_{p,j};L).
\end{equation}
A larger \(D_{\mathrm{J}}(L)\) indicates greater response diversity. We
additionally report the difference between OPSA and the base model as
\begin{equation}
\Delta D_{\mathrm{J}}(L)
=
D_{\mathrm{J}}^{\mathrm{OPSA}}(L)
-
D_{\mathrm{J}}^{\mathrm{Base}}(L).
\end{equation}

\section{Additional Results}

\subsection{Performance of Training on Code Generation Tasks}\label{app:code_opsa_train}
In this section, we evaluate OPSA in a broader domain to further assess its applicability. Unlike the main experiments, which focus primarily on mathematical reasoning, we study OPSA training on code-generation tasks.

\paragraph{Models and Datasets.}
We train OPSA on two model families with different reasoning modes, using \texttt{Qwen3-4B} in non-thinking mode and \texttt{Olmo3-7B-Instruct}. Training hyperparameters follow the settings described in Appendix~\ref{app:train_hyper}. We use Open-R1 Codeforces~\citep{penedo2025codeforces} as the training dataset and select the best checkpoint based on MBPP+ validation performance. Final in-domain evaluation is conducted on LiveCodeBench v6 and HumanEval+, while out-of-domain generalization is evaluated on the mathematical reasoning benchmark AIME24 and the general question-answering benchmark GPQA-Diamond~\citep{rein2023gpqa}.

\begin{table}[h]
    \centering
    \small
    \begin{tabular}{lccccc cc}
    \toprule
    \multirow{3}{*}{\textbf{Methods}} & \multicolumn{4}{c}{\textbf{Code (I.D.)}} & \multicolumn{3}{c}{\textbf{O.O.D.}}\\
    \cmidrule(lr){2-5} \cmidrule(lr){6-8}
    & \multicolumn{2}{c}{\textbf{LCB v6}}
    & \multicolumn{2}{c}{\textbf{HumanEval+}}
    & \multicolumn{2}{c}{\textbf{AIME24}}
    & \textbf{GPQA$_{D}$}\\
    & avg@32 & pass@32
    & avg@32 & pass@32
    & avg@32 & pass@32
    & avg@32 \\
    \midrule
    Qwen3-4B
    & 25.71 & 37.14 & 78.75 & 93.90 & 23.33 & 56.67 & 38.46 \\
    \rowcolor{cite!20}w/ OPSA & 37.57 & 52.00 & 85.79 & 97.56 & 28.96 & 66.67 & 41.04 \\
    \rowcolor{cite!20}$\Delta$ & \textcolor{myred}{+11.86} & \textcolor{myred}{+14.86} & \textcolor{myred}{+7.03} & \textcolor{myred}{+3.66} & \textcolor{myred}{+5.63} & \textcolor{myred}{+10.00} & \textcolor{myred}{+2.58} \\
    & $\uparrow$ 46.1\% & $\uparrow$ 40.0\% & $\uparrow$ 8.9\% & $\uparrow$ 3.9\% & $\uparrow$ 24.1\% & $\uparrow$ 17.6\% & $\uparrow$ 6.7\% \\
    \midrule
    Olmo3-7B-Instruct & 24.14 & 49.14 & 81.52 & 96.95 & 54.48 & 83.33 & 34.12 \\ 
    \rowcolor{cite!20} w/ OPSA & 32.09 & 55.43 & 83.62 & 98.17 & 56.25 & 86.67 & 35.66 \\
    \rowcolor{cite!20} $\Delta$ & \textcolor{myred}{+7.95} & \textcolor{myred}{+6.29} & \textcolor{myred}{+2.10} & \textcolor{myred}{+1.22} & \textcolor{myred}{+1.77} & \textcolor{myred}{+3.34} & \textcolor{myred}{+1.54} \\
    & $\uparrow$ 32.9\% & $\uparrow$ 12.8\% & $\uparrow$ 2.6\% & $\uparrow$ 1.3\% & $\uparrow$ 3.2\% & $\uparrow$ 4.0\% & $\uparrow$ 4.5\% \\
    \bottomrule
    \end{tabular}
    \caption{Performance of OPSA across different models on in-domain code generation and out-of-domain mathematical and general Q\&A tasks.}
    \label{tab:opsa_code}
\end{table}

\paragraph{Main Results.}
Table~\ref{tab:opsa_code} shows that OPSA consistently improves both in-domain code-generation performance and out-of-domain generalization across the two model families. On \texttt{Qwen3-4B}, OPSA improves Avg@32 by 11.86 points on LCB v6 and 7.03 points on HumanEval+, with corresponding Pass@32 gains of 14.86 and 3.66 points. Similar improvements are observed on \texttt{Olmo3-7B-Instruct}, where Avg@32 increases by 7.95 points on LCB v6 and 2.10 points on HumanEval+. These results demonstrate that OPSA remains effective when trained directly on code-generation tasks.
Interestingly, the gains also transfer beyond the training domain. On \texttt{Qwen3-4B}, OPSA improves AIME24 Avg@32 and Pass@32 by 5.63 and 10.00 points, respectively, and increases GPQA-Diamond Avg@32 by 2.58 points. \texttt{Olmo3-7B-Instruct} shows the same trend, with consistent improvements on both AIME24 and GPQA-Diamond. These results further demonstrate the robustness of OPSA across training domains, model families, and reasoning modes.

\subsection{Multi-Domain On-Policy Self-Adapatation}

\begin{table}[h]
    \centering
    \fontsize{8pt}{9pt}\selectfont
    \begin{tabular}{lc@{\hskip 4pt}c@{\hskip 5pt}c@{\hskip 3pt}c c@{\hskip 3pt}c@{\hskip 5pt}c@{\hskip 3pt}c@{\hskip 5pt}c@{\hskip 3pt}c}
    \toprule
    \multirow{3}{*}{\textbf{Methods}} 
    & \multicolumn{6}{c}{\textbf{Math}}
    & \multicolumn{4}{c}{\textbf{Code}}\\
    \cmidrule(lr){2-7} \cmidrule(lr){8-11}
    & \multicolumn{2}{c}{\textbf{AIME24}}
    & \multicolumn{2}{c}{\textbf{AIME25}}
    & \multicolumn{2}{c}{\textbf{HMMT25}}
    & \multicolumn{2}{c}{\textbf{LCB v6}}
    & \multicolumn{2}{c}{\textbf{HumanEval+}}\\
    & avg@32 & pass@32
    & avg@32 & pass@32
    & avg@32 & pass@32
    & avg@32 & pass@32
    & avg@32 & pass@32\\
    \midrule
    Qwen3-4B
    & 23.33 & 56.67
    & 20.52 & 56.67
    & 13.13 & 33.33
    & 25.71 & 37.14
    & 78.75 & 93.90 \\

    w/ OPSA$_\text{math}$
    & \underline{62.08} & \textbf{83.33}
    & \underline{58.44} & \textbf{83.33}
    & \textbf{37.40} & \underline{60.00}
    & 27.95 & 45.71
    & 83.35 & 96.34 \\

    w/ OPSA$_\text{code}$
    & 28.96 & \underline{66.67}
    & 22.92 & 63.33
    & 13.85 & 36.67
    & \textbf{37.57} & \underline{52.00}
    & \textbf{85.79} & \underline{97.56} \\

    w/ MD-OPSA
    & \textbf{66.77} & \textbf{83.33}
    & \textbf{60.83} & \underline{80.00}
    & \underline{36.88} & \textbf{70.00}
    & \underline{36.09} & \textbf{56.00}
    & \underline{84.96} & \textbf{98.78} \\
    \bottomrule
    \end{tabular}
    \caption{Performance comparison of single-domain OPSA (OPSA$_{{\text{math}}}$ and OPSA$_{{\text{code}}}$) and multi-domain OPSA (MD-OPSA) across math and code benchmarks. \textbf{Bold} and \underline{underlined} values indicate the best and second-best results, respectively. Training and evaluation use non-thinking mode.}
    \label{tab:md_opsa}
\end{table}

Multi-Teacher On-Policy Distillation (MOPD)~\citep{ma2026mopd} has been shown to enable effective capability merging during model post-training. In our previous experiments, we also found that OPSA trained on data from different domains achieves strong performance on both in-domain and out-of-domain evaluations. In particular, the substantial in-domain gains suggest that capability improvement may arise from reshaping the policy distribution through training on domain-specific data. This naturally raises a question: can we eliminate the domain-specific teacher models required by MOPD and instead achieve substantial multi-domain capability improvements by allowing the student policy to reshape its own distribution across data from different domains?

To investigate this question, we introduce \textbf{M}ulti-\textbf{D}omain \textbf{O}n-\textbf{P}olicy \textbf{S}elf-\textbf{A}daptation (\textbf{MD-OPSA}). MD-OPSA requires neither modifications to the training objective nor domain-specific routing or credit assignment. We simply randomly mix training data from different domains and optimize the policy using the same OPSA objective in Eq.~\ref{eq:opsa_loss}. More specifically, we randomly sample an equal number of math and code-generation prompts from DAPO-17k and Open-R1 Codeforces, respectively, which are the same training datasets used in our previous experiments. We train \texttt{Qwen3-4B} on the mixed dataset, with non-thinking mode enabled for both training rollouts and evaluation.

Table~\ref{tab:md_opsa} compares MD-OPSA with OPSA trained separately on math and code data. As expected, single-domain training primarily improves performance within its own domain: OPSA$_{\text{math}}$ yields substantial gains on mathematical reasoning, while OPSA$_{\text{code}}$ performs best on several code-generation metrics. In contrast, MD-OPSA successfully combines the benefits of both domains, achieving strong improvements across all math and code benchmarks without domain-specific teachers, routing, or credit assignment. Notably, MD-OPSA even surpasses the corresponding single-domain specialists on several metrics, including Avg@32 on AIME24 and AIME25 and Pass@32 on LCB v6 and HumanEval+. On the remaining benchmarks, its performance remains close to the best single-domain model. These results suggest that simply mixing multi-domain data allows OPSA to reshape the student policy toward capabilities from multiple domains with little apparent interference, providing a simple teacher-free alternative for capability merging. Interestingly, the effects of distribution reshaping across domains may not be entirely orthogonal: adaptation on one domain can also provide transferable benefits to others, which may help explain the strong cross-domain performance of MD-OPSA.

\subsection{Comparison to Negative Sample Reinforcement}
\begin{table}[h]
    \centering
    \small
    \begin{tabular}{l@{\hskip 7pt}c@{\hskip 5pt}cc@{\hskip 5pt}cc@{\hskip 5pt}cc@{\hskip 5pt}c}
    \toprule
    \multirow{2}{*}{\textbf{Methods}}
    & \multicolumn{2}{c}{\textbf{AIME24}}
    & \multicolumn{2}{c}{\textbf{AIME25}}
    & \multicolumn{2}{c}{\textbf{HMMT25}}
    & \multicolumn{2}{c}{\textbf{Average}} \\
    & avg@32 & pass@32
    & avg@32 & pass@32
    & avg@32 & pass@32
    & avg@32 & pass@32 \\
    \midrule
    Qwen3-1.7B
    & 13.44 & 40.00
    & 9.69 & 30.00
    & 5.73 & 23.33
    & 9.62 & 31.11 \\
    + GRPO
    & 33.96 & 70.00
    & 25.31 & 50.00
    & 15.10 & 43.33
    & 24.79 & 54.44 \\
    + NSR
    & 32.08 & 73.33
    & 24.17 & 60.00
    & 16.15 & 43.33
    & 24.13 & 58.89 \\
    \rowcolor{cite!20}
    + OPSA
    & 48.85 & 80.00
    & 35.31 & 66.67
    & 23.33 & 50.00
    & 35.83 & 65.56 \\
    \bottomrule
    \end{tabular}
    \caption{Comparison of different on-policy RL methods on
    \texttt{Qwen3-1.7B}.}
    \label{tab:nsr}
\end{table}

NSR~\citep{zhu2026surprising} decomposes RLVR trajectories into correct and incorrect samples and performs trajectory-level reinforcement learning with negative advantages. Specifically, it assigns a fixed negative advantage to every token in an incorrect trajectory, thereby improving exploration while maintaining stable policy entropy during training. The key distinction between OPSA and NSR is that OPSA requires no supervisory signal, including verifiable rewards. Instead, it assigns token-level negative advantages to low-log-probability tokens across all trajectories and dynamically modulates these advantages according to token entropy, enabling substantially finer-grained optimization.
Combined with our analysis in Section~\ref{sec:4.3}, these results suggest that OPSA suppresses low-probability tail tokens and redistributes their probability mass more evenly among high-probability head tokens. This mechanism preserves accuracy at low-entropy positions while enhancing exploration at high-entropy fork tokens. As shown in Table~\ref{tab:nsr}, OPSA consistently outperforms both GRPO and NSR in single-sample accuracy and pass@$k$, validating the effectiveness of our approach. The results further demonstrate that applying negative advantages to low-probability tokens even within correct trajectories is important for improving model performance.

\subsection{Low Entropy Can Still Support Effective Exploration}
\begin{figure}[h]
    \centering
    \includegraphics[width=1\linewidth]{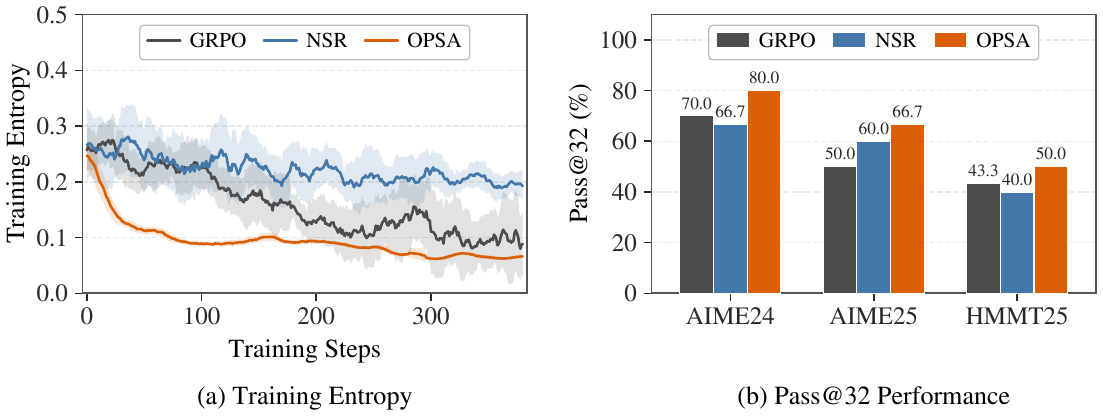}
    \caption{Training dynamics of different methods on \texttt{Qwen3-1.7B} and pass@32 performance.}
    \label{fig:entropy}
\end{figure}

As shown in Figure~\ref{fig:entropy}, OPSA exhibits substantially lower training entropy than GRPO and NSR while consistently achieving higher pass@32 on AIME24, AIME25, and HMMT25. This result demonstrates that aggregate entropy alone is not a reliable indicator of a model's exploration ability. Effective exploration depends more critically on how uncertainty is allocated across token positions. At high-entropy fork tokens, OPSA maintains a relatively balanced probability distribution among plausible head tokens, enabling the model to explore multiple promising reasoning branches. At low-entropy positions, it further concentrates probability mass on high-confidence tokens, improving sampling accuracy and preventing trajectories from entering implausible, low-confidence branches. Consequently, OPSA achieves lower overall entropy without sacrificing diversity and meaningful exploration at critical decision points, consistent with our analysis in Section~\ref{sec:5.3}.

\subsection{Performance Comparison Under Similar Token Budgets}

\begin{table}[h]
    \small
    \centering
    \begin{tabular}{lcc}
    \toprule 
        \textbf{Methods} & \textbf{Token Counts}  & \textbf{Avg@32} \\
        \midrule
        Qwen3-1.7B  & 4457  & 13.44 \\
        + GRPO  & 19108  & 33.96 \\
        + GRPO (``wait")  & 23261 & 32.81 \\
        + OPD & 15286  & 32.08 \\
        + OPD (``wait") & 23472  & 31.67 \\
        + OPSA & 23205 & 48.85 \\
        \bottomrule
        
    \end{tabular}
    \caption{AIME24 performance of different RL strategy on similar inference token budgets.}
    \label{tab:control_token}
\end{table}

To verify that OPSA’s gains are not merely a consequence of generating longer responses, we construct a token-budget-matched inference control for the GRPO and OPD baselines. Specifically, we first generate a baseline response, remove its terminal \verb|\boxed{}| answer, append a single ``wait'' token to the remaining reasoning trace, and then resume decoding. A minimum generation-length constraint is applied so that the resulting responses approximately match the average response length of the OPSA-trained model. We evaluate the first complete \verb|\boxed{}| answer produced after reaching the target token budget while keeping all other decoding settings unchanged. As shown in Table~\ref{tab:control_token}, allocating additional inference-time tokens does not improve the baselines, and OPSA continues to substantially outperform both GRPO and OPD under comparable response lengths. These results indicate that OPSA’s improvements cannot be explained by response length alone. Instead, they are consistent with OPSA reshaping the policy distribution to improve precision at low-entropy positions and encourage exploration at high-entropy positions, thereby steering generation toward more effective reflective reasoning branches.

\subsection{OPSA as a Cold-Starting for GRPO Training}

\begin{figure}[h]
    \centering
    \includegraphics[width=0.8\linewidth]{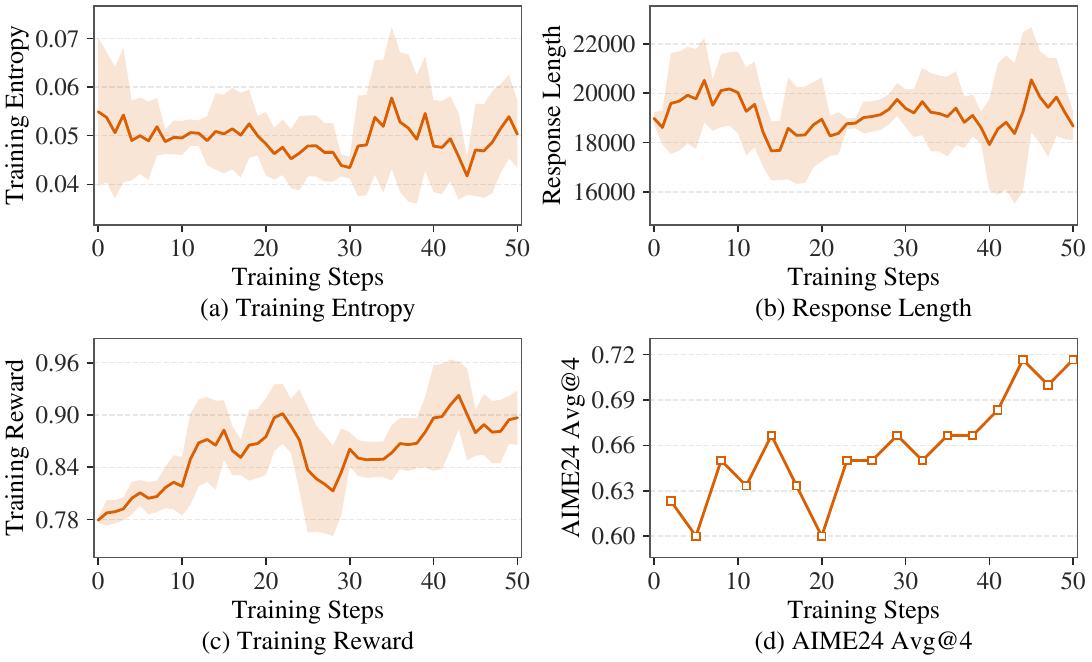}
    \caption{Training dynamics of GRPO training started from the OPSA cold-starting 4B model.}
    \label{fig:opsa_coldstart}
\end{figure}
OPSA demonstrates strong effectiveness during post-training, improving both accuracy and Pass@$k$ while preserving sampling diversity. In this section, we investigate whether OPSA, as an external-supervision-free method, can serve as a ``free-lunch'' cold start before further RL training and enable additional performance gains. We conduct experiments on \texttt{Qwen3-4B}, using an OPSA checkpoint as the initialization and further training it with GRPO on DAPO-17k. Figure~\ref{fig:opsa_coldstart} reports the resulting training dynamics. On the validation set, Avg@4 continues to improve steadily, increasing by approximately 9 points after 40 training steps, while the training curve remains smooth without signs of collapse. These results suggest that OPSA can provide an effective cold start for subsequent RL training, further extending the model’s capabilities.

\subsection{OPSA Elicits Reasoning Capabilities from an Instruct Model}
\begin{figure}[h]
    \centering
    \includegraphics[width=1.0\linewidth]{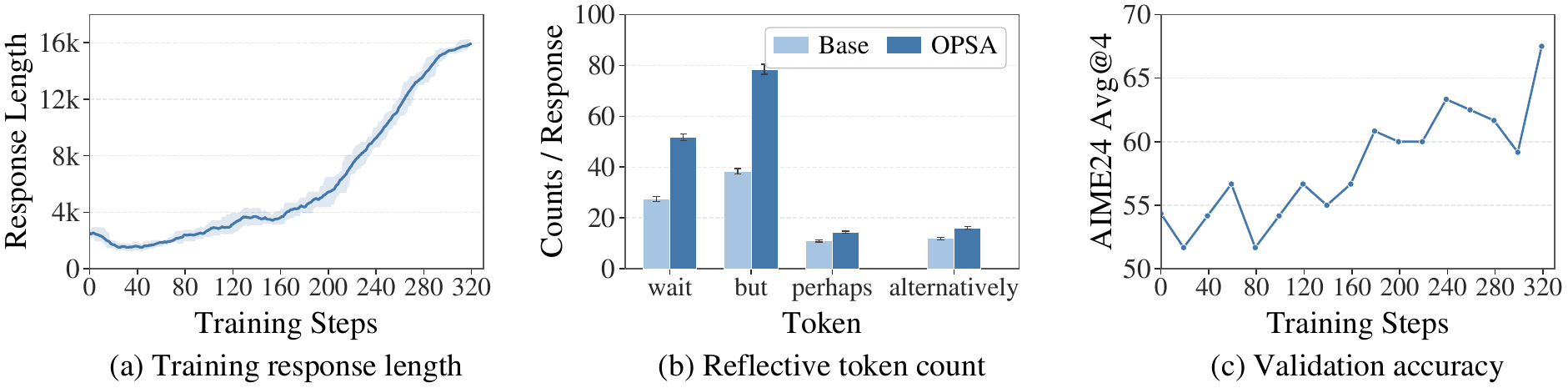}
    \caption{Training dynamics of OPSA on Olmo3-7B-Instruct}
    \label{fig:olmo_dynamic}
\end{figure}

Our previous analysis in Section~\ref{sec:5.2} suggests that OPSA improves reasoning by steering the policy toward longer and more reflective reasoning branches. Since Qwen3 is a hybrid model with an explicit thinking mode, it remains unclear whether this behavior simply reflects the recovery of reasoning patterns already encoded in the model. To examine whether OPSA can induce similar behavior beyond hybrid models, we further apply OPSA to \texttt{Olmo3-7B-Instruct}. As shown in Fig.~\ref{fig:olmo_dynamic}, OPSA steadily increases response length together with validation accuracy, while substantially increasing the frequency of reflective tokens such as ``wait'' and ``but''. These results suggest that OPSA does not merely recover a pre-existing thinking mode. Instead, by reshaping the policy distribution at high-entropy fork positions, it increases the likelihood of exploring reflective reasoning branches and can elicit stronger reasoning behavior even from an instruct model.

\subsection{Tuning Inference-Time Sampling Hyperparameters Can't Replace OPSA}
\begin{figure}[h]
    \centering
    \includegraphics[width=1.0\linewidth]{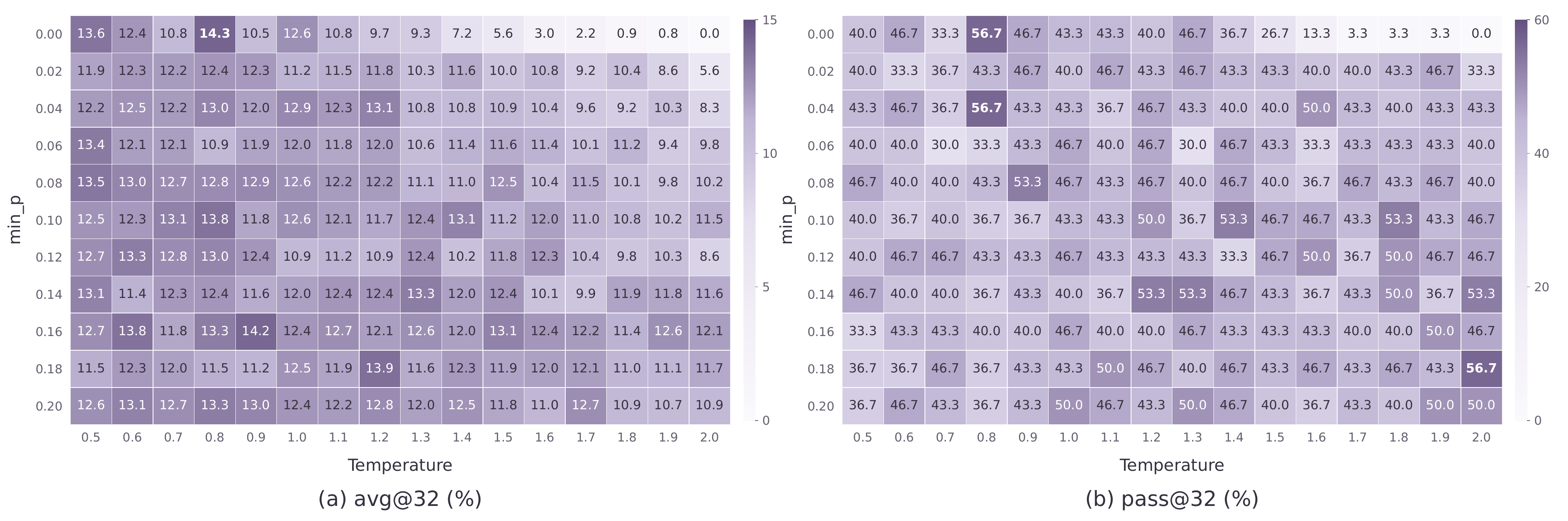}
    \caption{Results on AIME24 with \texttt{Qwen3-1.7B} in non-thinking mode under different min-p and temperature settings: (a) Avg@32 and (b) Pass@32. All values are percentages, and \textbf{bold} entries indicate the best result in each panel.}
    \label{fig:inf_tune}
\end{figure}

Although tuning inference-time sampling parameters can also suppress low-probability tail tokens and redistribute sampling probability among head tokens, Fig.~\ref{fig:inf_tune} shows that it cannot reproduce the gains of OPSA. We attribute this gap to two key differences. First, inference-time sampling applies a fixed transformation to the base policy at every decoding step without changing the underlying policy itself. However, the desired amount of tail suppression and head-token redistribution can vary substantially across token positions and reasoning states, making it difficult to determine a single optimal temperature or min-p value in advance. In contrast, OPSA adapts the magnitude of its updates according to the policy’s token-level uncertainty, enabling state-dependent distribution reshaping through training rather than relying on a globally fixed sampling rule. Second, OPSA performs this reshaping iteratively: each policy update changes the distribution from which subsequent trajectories are sampled, exposing new reasoning branches that can be further optimized. Such parameter updates can also generalize across contexts and alter the modes of the token distribution, while inference-time sampling continues to operate on the same underlying base-policy modes. Thus, although both approaches can suppress tail tokens, OPSA learns a new reasoning policy rather than merely sampling differently from the original one.

\end{document}